\documentclass[10pt,twocolumn,letterpaper]{article}

\usepackage{iccv}
\usepackage{times}
\usepackage{epsfig}
\usepackage{graphicx}
\usepackage{amsmath}
\usepackage{amssymb}
\usepackage[accsupp]{axessibility}  

\usepackage[pagebackref=true,breaklinks=true,letterpaper=true,colorlinks,bookmarks=false]{hyperref}

\iccvfinalcopy 
\usepackage[capitalize]{cleveref}
\crefname{section}{Sec.}{Secs.}
\Crefname{section}{Section}{Sections}
\Crefname{table}{Table}{Tables}
\crefname{table}{Tab.}{Tabs.}

\usepackage{float}
\usepackage{subcaption}
\usepackage{enumitem} 
\usepackage{booktabs}
\usepackage{multirow}
\usepackage{rotating}
\usepackage{adjustbox}
\def\eg{\emph{e.g}\onedot} 

\def\ie{\emph{i.e}\onedot} 

\usepackage{mathtools}
\usepackage{textcomp}
\usepackage{gensymb}
\usepackage[dvipsnames]{xcolor}

\newcommand{\myparagraph}[1]{\vspace{4pt}\noindent\textbf{#1}}
\newcommand\our{EigenPlaces}

\def\iccvPaperID{9537} 

\ificcvfinal\fi

\begin{document}

\title{EigenPlaces: Training Viewpoint Robust Models for Visual Place Recognition}

\author{Gabriele Berton*$^{1}$
\quad
Gabriele Trivigno*$^{1}$
\quad
Barbara Caputo$^{1}$
\quad
Carlo Masone$^{1}$\\
$^{1}$ Politecnico di Torino\\
{\tt\small \{gabriele.berton, gabriele.trivigno, barbara.caputo, carlo.masone\}@polito.it}\\
}

\maketitle
\ificcvfinal\thispagestyle{empty}\fi

\begin{abstract}
Visual Place Recognition is a task that aims to predict the place of an image (called query) based solely on its visual features.
This is typically done through image retrieval, where the query is matched to the most similar images from a large database of geotagged photos, using learned global descriptors.
A major challenge in this task is recognizing places seen from different viewpoints. To overcome this limitation, we propose a new method, called EigenPlaces, to train our neural network on images from different point of views, which embeds viewpoint robustness into the learned global descriptors. The underlying idea is to cluster the training data so as to explicitly present the model with different views of the same points of interest. The selection of this points of interest is done without the need for extra supervision.
We then present experiments on the most comprehensive set of datasets in literature, finding that EigenPlaces is able to outperform previous state of the art on the majority of datasets, while requiring 60\% less GPU memory for training and using 50\% smaller descriptors.
The code and trained models for EigenPlaces are available at {\small{\url{https://github.com/gmberton/EigenPlaces}}}, while results with any other baseline can be computed with the codebase at 
{\small{\url{https://github.com/gmberton/auto_VPR}}}.
\end{abstract}


\begin{figure}
    \begin{center}
    \begin{minipage}{.48\textwidth}
        \includegraphics[width=\textwidth]{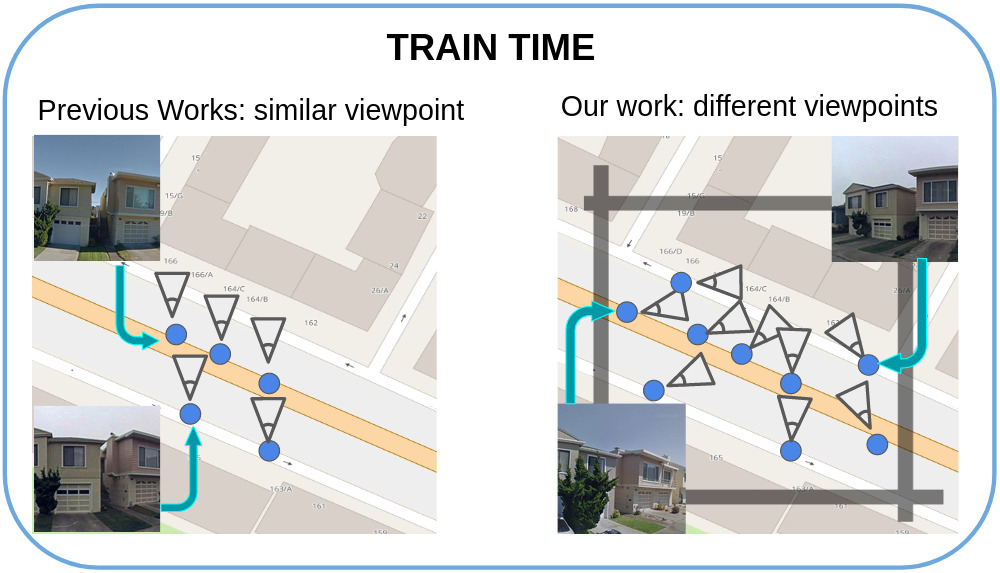}
    \end{minipage}
    \begin{minipage}{.48\textwidth}
    \includegraphics[width=\linewidth]{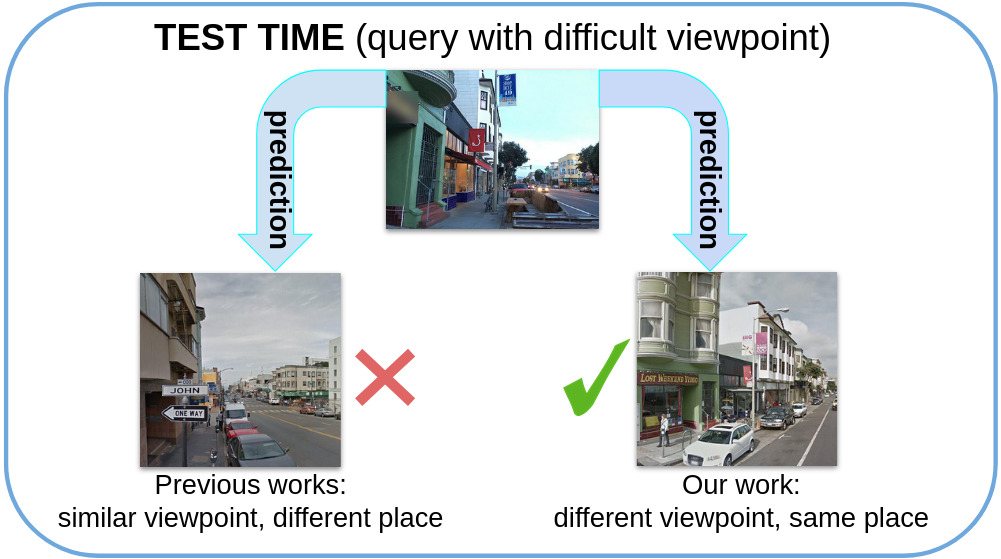}
    \end{minipage}
    \end{center}
    \caption{
    Most previous works \cite{Arandjelovic_2018_netvlad, Liu_2019_sare, Kim_2017_crn, Zhang_2021_gated_netvlad} train their models through metric learning, using as positive the most similar image to the query, which naturally would have same or similar orientation to the query. Other works split the dataset in classes, with images within a class having similar \cite{Alibey_2022_gsvcities, Alibey_2023_mixvpr} or exactly the same \cite{Berton_2022_cosPlace} orientation.
    EigenPlaces goes against this trend, creating classes in which all images are oriented towards the same point, leading to viewpoint robust models able to correctly localize highly challenging queries, for example the ones collected from a sidewalk.}
    \label{fig:teaser}
\end{figure}

\section{Introduction}
\label{sec:introduction}
Visual Place Recognition (VPR) is a task that aims to predict the place where a photo (\ie query) was taken, quickly and accurately, based solely on its visual features. This is typically done with an image retrieval approach~\cite{Torii_2015_pitts250k, Chen_2017a,  Kim_2017_crn, Chen_2017b, Arandjelovic_2018_netvlad, Chen_2018, Torii_2018_tokyo247, Hausler_2019, Garg_2019,  Liu_2019_sare, Doan_2019,  Ge_2020_sfrs, Khaliq_2020,  Warburg_2020_msls, Torii_2021_r_sf, Hausler_2021_patch_netvlad, Ibrahimi_2021_insideout_vpr, Zaffar_2021_vprbench, Mereu_2022_seqvlad, Alibey_2022_gsvcities, Berton_2022_cosPlace, Berton_2022_benchmark, Alibey_2023_mixvpr, Leyvavallina_2021_gcl}: first, a deep neural network is used to extract global descriptors from the query and from a database of geo-referenced images; then, a nearest neighbor search is performed in this features space \cite{Arandjelovic_2018_netvlad, Liu_2019_sare, Kim_2017_crn, Ge_2020_sfrs, Berton_2022_cosPlace, Alibey_2023_mixvpr, Alibey_2022_gsvcities}.
While such approaches have shown great potential in partially solving known problems such as scalability (by using ever more compact descriptors \cite{Zhu_2018_apanet, Berton_2022_cosPlace, Alibey_2023_mixvpr}) and illumination changes (through the synthetic generation of night images \cite{Berton_2021_svox, Porav_2018_gan_for_geoloc, Asha_2019_todaygan} or strong data augmentation \cite{Ge_2020_sfrs}), 
recognizing images under heavy viewpoint shifts is still an open challenge.
A popular strategy to handle this problem is to follow up the similarity search performed on global feature descriptors with a post-processing phase that re-ranks the retrieved results using either spatial verification \cite{Cao_2020_delg, Liu_2020_densernet, Hausler_2021_patch_netvlad} or matching densely extracted local features descriptors \cite{Berton_2021_geowarp}.
However, these post-processing methods are useless if the similarity search is not able to retrieve at least one matching result from the database. Moreover, these techniques are also costly, given that the local feature matching must be performed for each retrieved results, thus the number of candidates to be re-ranked is usually orders of magnitude smaller than the database \cite{Cao_2020_delg, Berton_2021_geowarp, Hausler_2021_patch_netvlad, Liu_2020_densernet}.

\begin{figure}
        \begin{center}
    \includegraphics[width=0.4\textwidth]{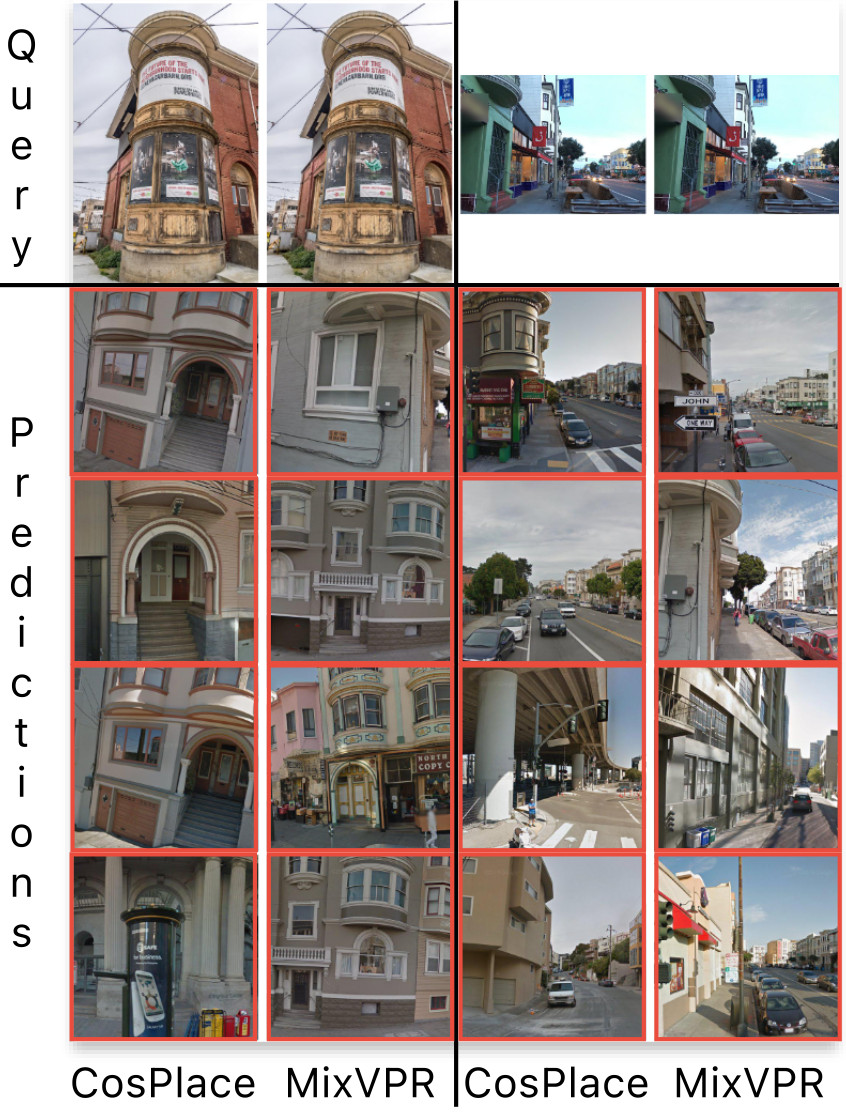}
    \end{center}
    \caption{\textbf{Examples of 2 challenging queries}
    (top row)
    which present heavy viewpoint changes with respect to the database.
    The first and third columns are matches to the two query obtained with CosPlace \cite{Berton_2022_cosPlace}, while the second and fourth column are matches obtained with MixVPR \cite{Alibey_2023_mixvpr}.}
    \label{fig:difficult_queries}
\end{figure}

Even so, we can observe that when the queries to be recognized present heavy viewpoint shifts with respect to the database images, state-of-the-art retrieval architectures fail to find any relevant result in the highest ranked candidates (see \cref{fig:difficult_queries}). 
In view of these considerations, we argue that it is necessary to improve the robustness to viewpoint shifts already at the retrieval stage, by teaching the network to extract global descriptor extractor that are more invariant to perspective changes.
To achieve this goal we propose EigenPlaces, a new training paradigm that clusters the training data into classes so that each class contains only multiple viewpoints depicting the same scene. This forces the model to learn global descriptors that are robust to viewpoint shifts (see \cref{fig:teaser}).
This is done by estimating the presence of \textit{places} (such as building facades) based solely on the geographical distribution of training data, by splitting the training dataset in classes and finding the geographical principal components within each given class.

To empirically show the soundness of our method, we run a benchmark on the largest number of VPR datasets ever. The results show that a model trained with EigenPlaces is able to outperform previous SOTA on numerous datasets, while using 50\% smaller descriptors and requiring 60\% less GPU memory for training.

\myparagraph{Our contributions} are summarized as follows:
\begin{itemize}
    \item we propose EigenPlaces, a novel training protocol whose ultimate goal is to render the model robust to viewpoint changes that it may encounter at test time;
    \item we perform a rigorous VPR benchmark on the most complete set of datasets in literature, to highlight not only the strengths but also the weaknesses of EigenPlaces and its predecessors;
    \item while our exploration shows that there is no one-win-all solution on every scenario, we note that EigenPlaces outperforms previous state of the art on a large number of datasets, while needing 60\% less GPU memory to train and using 50\% more compact descriptors.
\end{itemize}


The code and trained models for EigenPlaces are available at {\small{\url{https://github.com/gmberton/EigenPlaces}}}.

We also created and released a codebase to run experiments with a number of trained models (namely NetVLAD, SFRS, CosPlace, Conv-AP, MixVPR and EigenPlaces), which automatically downloads each model's weights from their official repository, to be able to run experiments within a fair and standardized framework.
The codebase is available at {\small{\url{https://github.com/gmberton/auto_VPR}}}.



\section{Related Work}
\label{sec:related_work}
\myparagraph{Visual place recognition.} Most early works on VPR focused on matching queries to their database counterparts through the use of local features \cite{Torii_2015}, with methods such as SIFT \cite{Lowe_2004_sift}, SURF \cite{Bay_2008_surf} and RootSIFT \cite{Arandjelovic_2012_rootSift} dominating the pre-deep learning landscape, although the use of global or patch features has also been investigated \cite{Oliva_2006_gist}.
With the advent of deep learning, \cite{Babenko_2014_neural_codes} found that features extracted with a CNN trained for classification can be successfully used for landmark retrieval.
This inspired a number of following works, which used the same concept to extract global learned features with a number of pooling layers \cite{Razavian_2015_mac, Tolias_2016_rmac,Radenovic_2019_gem}.
To ensure that the model learns to extract specific features for urban VPR, 
Arandjelovic \etal \cite{Arandjelovic_2018_netvlad} proposed to train it on a dataset of StreetView images, while enhancing the CNN with a novel layer, named NetVLAD, that encodes 3D features maps to a highly informative vector.
A number of subsequent works built on top of NetVLAD, enhancing it with an attention module~ \cite{Kim_2017_crn}, a novel loss~\cite{Liu_2019_sare}, or a  self-supervised strategy to crop training images~\cite{Ge_2020_sfrs}.
A different training strategy was introduced by CosPlace \cite{Berton_2022_cosPlace}, which takes inspiration from the face recognition literature \cite{Liu_2017_sphereface, Wang_2018_cosFace, Deng_2019_arcFace} to train the model through a classification task, and then uses the same learned features to perform the retrieval.
Recently, some works have also shown dramatic improvements training on the large-scale dataset of GSV-cities \cite{Alibey_2022_gsvcities} using a Multi-Similarity loss \cite{Wang_2019_multi_similarity_loss}, and processing the high-level features extracted by a CNN with a newly proposed Conv-AP layer \cite{Alibey_2022_gsvcities} or a Feature-Mixer \cite{Alibey_2023_mixvpr}.

Despite the strides made by all these methods over the last years, none of them have explicitly addressed the viewpoint-invariance problem for visual place recognition.
In particular, NetVLAD and its derivatives base their learning protocol on the use of the most similar database images to the query as positives, which are likely to share the same (or similar) viewpoint. 
MixVPR \cite{Alibey_2023_mixvpr} uses a set of pre-defined images, grouped in classes, all of which share a similar viewpoint. Finally, in CosPlace all images in a class have exactly the same orientation by design.

\myparagraph{Viewpoint invariant matching.}
There is also a separate body of literature dedicated to the refinement of a shortlist of candidates provided by an image retrieval module, by matching local features. Popular examples are SuperGlue \cite{Sarlin_2020_superglue}, DELG \cite{Cao_2020_delg}, LoFTR \cite{Sun_2021_loftr}, Patch-NetVLAD \cite{Hausler_2021_patch_netvlad}, GeoWarp \cite{Berton_2021_geowarp}, and others \cite{Wang_2022_TransVPR, Fuwen_2021_reranking_transformers}. These approaches are based on the premise that by leveraging local features associated to detected keypoints, it is possible to match landmarks even from different viewpoints. 

Although these methods can mitigate the viewpoint shift problem, they all rely on the assumption that the shortlist of candidates resulting from retrieval stage contains at least a positive match. Our method is oriented towards improving the robustness of the retrieval stage, thus it is complementary to all these approaches. Moreover, since these techniques are computationally expensive, being able to retrieve a positive result is crucial to their applicability.

\section{Method}
\label{sec:method}
Despite recent advances in the literature, substantial changes in viewpoint still represent a challenge even for modern SOTAs (see some examples in \cref{fig:difficult_queries}).
In practice this kind of distribution shift is very common, because the database images for retrieval are usually collected via car-mounted cameras \cite{Torii_2015_pitts250k, Torii_2018_tokyo247, Maddern_2017_robotCar, Berton_2021_svox, Alibey_2022_gsvcities, Milford_2008_st_lucia, Chen_2011_san_francisco, Warburg_2020_msls}, whereas the queries may come from different sources (\eg photos taken by smartphones \cite{Torii_2018_tokyo247, Chen_2011_san_francisco, Berton_2022_cosPlace}) and can present a substantial variability in terms of viewpoint. 

Knowing the positions of all \textit{places} or \textit{points of interest} (\eg a building facade or architectural landmark) within the map, a straightforward approach to mitigate the issue could be to find all images of our train set that face towards them (\ie from all viewpoint), and minimize a loss that pulls together features representing the same place.
However, annotating the positions of all buildings in a city can be a challenging and expensive task, especially in a high density scenario, which would limit the scalability and practicality of a geolocalization system.
To overcome this impediment, in this work we introduce a training algorithm that is able to automatically obtain different views of a given \textit{place}, \ie images that look at the same scene from different angles. Our novel method estimates the direction of a road using only the images' coordinates, and builds on the premise that points of interest lie on the side of the road.

In the following sections, we describe how we use EigenPlaces to train our networks:
\begin{itemize}[noitemsep,topsep=1pt]
    \item in \cref{sec:dataset_partitioning} we explain how we split a dense dataset in non-overlapping cells, avoiding the risk of having images of the same place contained in different classes;
    \item in \cref{sec:eigenplaces} we present how EigenPlaces selects a subset of images within a given cell, representing different views of the same place;
    \item in \cref{sec:training} we show the loss used to train the model using the selected images.
\end{itemize}


\subsection{Map Partition}
\label{sec:dataset_partitioning}
As a first preparatory step for {\our}, we divide the map in $M \times M$ cells, with $M = 15 ~meters$.
Next we group the cells in subsets, ensuring that within a single subset there are no neighboring cells. This guarantees that images within two cells of the same subset have no visual overlap, and thus cells within a subset can be later treated as classes for a classification task.
To this end, we take inspiration from CosPlace \cite{Berton_2022_cosPlace}, which has recently shown that a similar split can lead to good results on VPR.
To build the subsets, we therefore take only one every $N$ cells both in the latitudinal and longitudinal directions.
Thus, we consider only $1/N^2$ of the cells at a time and during training we shift the set of cells after each epoch.
Although this partitioning strategy is somewhat related to CosPlace's, our design relies only on the position of each image, and it does not entail the use of their orientation. Moreover, the rationale with which the classes are constructed from the cells is fundamentally different, and it is detailed in the next section.


\subsection{EigenPlaces}
\label{sec:eigenplaces}

\begin{figure}
    \begin{center}
    \includegraphics[width=0.8\columnwidth]{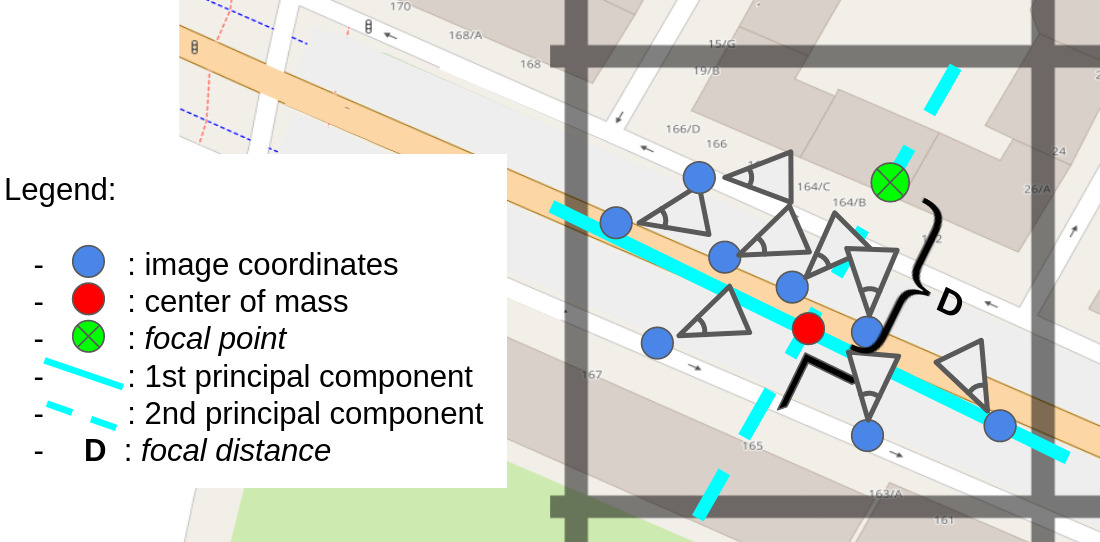}
    \end{center}
    \caption{\textbf{Sketch of EigenPlaces's principle.}
    For a given cell, the first and second principal components are derived from the images' positions.
    The first is an estimate of a road, whereas along the second we can find \textit{points of interest}, like facades.
    We choose a \textit{focal point} on the second principal components, and then we find the images pointing towards it to represent different views of the same place. Using these different views to train a model endows it with robustness to viewpoint shifts.
    Note how the \textit{focal distance} D defines the \textit{focal point}: a $D=0$ would lead to all the images pointing towards the center of mass of the images (\ie images pointing towards each other), whereas a $D \rightarrow \infty$ makes the images pointing towards an infinitely far \textit{focal point}, meaning that each image would have the same direction.
    }
    \label{fig:eigen_map}
\end{figure}

Given a cell from the map partition, all the images therein represent the same location and may be naively considered as a unique class by a classifier. However, the images may be taken by cameras pointing in different directions, so they may observe different scenes.
Thus, a model may struggle to learn a coherent representation for them all.
In CosPlace, this problem is solved by dividing a cell in multiple classes, where each class contains all the images oriented in the same geographical direction. 
Our idea is instead to select in a class all the images that look at a same place, from different perspectives.
To implement this idea without any extra supervision, we leverage the prior knowledge that database images are commonly collected by cameras mounted on cars \cite{Torii_2015_pitts250k, Torii_2018_tokyo247, Maddern_2017_robotCar, Berton_2021_svox, Alibey_2022_gsvcities, Milford_2008_st_lucia, Chen_2011_san_francisco, Warburg_2020_msls}, \ie they are aligned along roads.
Following this idea, we can assume that we can find distinctive \textit{points of interest} (\eg building facades) by looking at the side of the roads. 

To explain how we select these images, let us consider without lack of generality the $i$-th cell. 
Furthermore, let us denote as $X_i \in \mathbb{R}^{p \times 2}$, the matrix containing the UTM coordinates (east, north) of the $p$ images in the cell. 
Then, we compute the Singular Value Decomposition (SVD)  of the centered matrix $\widehat{X_i} = X_i - \text{E}[X_i]$.
Since $\widehat{X_i}$ is real-valued, we are guaranteed that the set of singular vectors obtained from the decomposition exists and is a orthonormal basis that can be ordered with respect to a set of non-negative singular values. Moreover, since the matrix is centered, the singular vectors are also the eigenvectors of the correlation matrix, \ie the principal components.\\
The first eigenvector represents the direction of maximum variability in our data. As discussed before, this direction likely corresponds with the road traveled by the vehicle that collected the images.
Therefore, the second eigenvector, perpendicular to the first one (and thus to the road), is likely directed to the side of the road.
Consequently, we can define a focal point on the second principal component, likely towards a building's facade.
Formally, we define the \textit{focal point} $c_i$ as :
\begin{equation}
    c_i = \text{E}[X_i] + D \times V_1
    \label{eq:conv_point}
\end{equation}
where $V_1$ is the second principal component obtained from the SVD decomposition and $D$ is a \textit{focal distance} from the center of mass which determines the exact location of the \textit{focal point} (see \cref{fig:eigen_map}).
Finally, within the given cell $i$, the images facing the \textit{focal point} $c_i$ are grouped in a single class and used effectively for training.
Note the importance of the \textit{focal distance} D for the construction of classes: when $D \rightarrow \infty$ the method selects all the images oriented in the same geographical direction (same orientation), whereas when $D\rightarrow 0$ the \textit{focal point} gets closer to the mean of the images position and the method selects images facing in opposite directions.

This method assumes that the images available in the database are collected looking at all sides of the vehicle, and in particular towards the side of the street. However, this is not always the case and many VPR datasets: for example, the datasets built with autonomous driving applications in mind only contain images collected from a front facing camera (St Lucia \cite{Milford_2008_st_lucia}, MSLS \cite{Warburg_2020_msls}, SVOX \cite{Berton_2021_svox}, RobotCar \cite{Maddern_2017_robotCar}).
In order to handle these cases, we repeat the same procedure to generate a second \textit{focal point} along the first right eigenvector (the one aligned with the direction of the road) and create a second class from the front-facing images.

Although this method is built on the intuition that the images in a cell are likely aligned in a straight line along a road, this is certainly not true in general. For instance, at crossroads the images are distributed along multiple directions. In such cases, the eigenvectors obtained from the SVD are not aligned/orthogonal with the road, and the points of interest may end up not on buildings but somewhere else. Nevertheless, this does not detract from the method, as the goal is to feed the model with images looking at the same point from different perspectives. On the contrary, having some variability in the  data so that not all points of interest are on buildings is helpful to make the model more robust.


\subsection{Training}
\label{sec:training}

\begin{figure}
    \begin{center}
    \includegraphics[width=0.6\columnwidth]{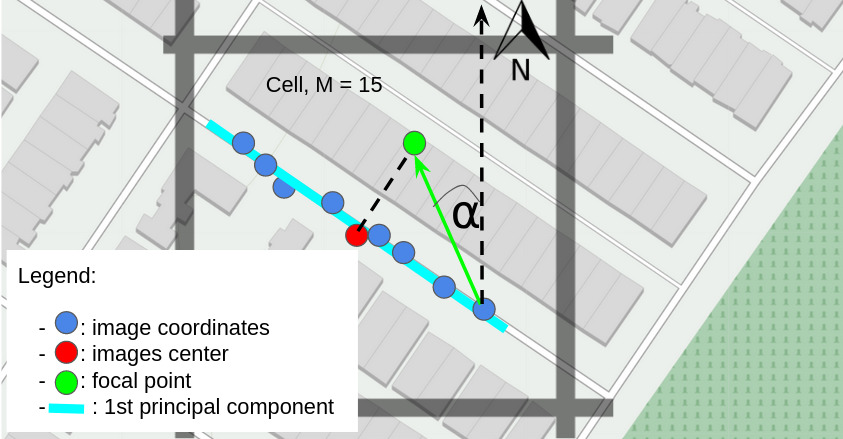}
    \end{center}
    \caption{\textbf{Construction of a class in Eigenplaces.}
    To select a relevant subset of images, for each image we compute the angle $\alpha$ as shown in the figure. We then select only the images whose orientation is close to $\alpha$.
    }
    \label{fig:train_cell}
\end{figure}

We now describe how to select images with changing viewpoints, once for a given class $i$ we have obtained its \textit{focal point} $c_i$ according to \cref{eq:conv_point}. Given an image $j$ and its UTM coordinates $x_j = (e_j, n_j)$:
\begin{equation}
\begin{aligned}
    c_i =& (e_c, n_c) \\
    \Delta_{e_j} = e_c - e_j, \, &
    \Delta_{n_j} = n_c - n_j \\
    \alpha_j =& \arctan(\frac{\Delta_{e_j}}{\Delta_{n_j}}) 
\end{aligned}
\label{eq:train_cell}
\end{equation}

In practice, we select images whose orientation is closest to $\alpha_j $. This computation is also exemplified in \cref{fig:train_cell}. The angle $\alpha_j$ depends on the relative positions of the image and the \textit{focal point}, and it represents the \textit{orientation}, \ie the deviation \wrt the north axis. The sign of $\alpha_j$ will vary if the image lies on the left of the second principal component.
Note that $\alpha_j$ varies for each of the images, and this is a key element in our approach, that allows to have the same \textit{place} depicted from different viewpoints.

Once the dataset is split in classes, and a number of images are selected for each class, we can use such data to train in an end-to-end fashion a deep neural network.
To this end, we use a Large Margin Cosine Loss (CosFace) \cite{Wang_2018_cosFace}, which has been shown to produce strong results in VPR \cite{Berton_2022_cosPlace}.
The CosFace layer is defined by a single fully connected with weights matrix $W^{lat}$, and the loss is computed as follows:
\begin{equation}
    \mathcal{L}_\textit{lat} = 
    \frac{1}{N} \sum_{i}{- log \frac{e^{s(\cos(\theta_{y_i}) - m)} }
         {e^{s(\cos(\theta_{y_i}) - m)} + \sum_{i \neq j}{e^{s\cos \theta_j}}}
    }
\end{equation}
subject to 
\begin{equation}
\begin{aligned}
    \cos \theta_j &= W_{lat_j}^T x_i \\
    W_{lat} = \frac{W^*}{||W^*||}, \quad &x = \frac{x^*}{||x^*||}
\end{aligned}
\end{equation}

\begin{figure}
    \begin{center}
    \includegraphics[width=0.19\textwidth]{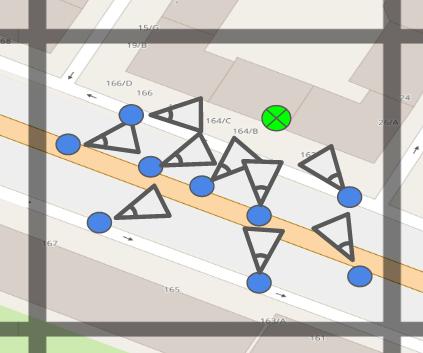}
    \includegraphics[width=0.19\textwidth]{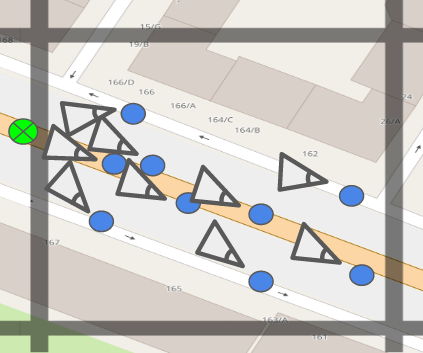}
    \end{center}
    \caption{
    \textbf{Lateral vs Frontal Loss.} The image on the left shows how the views are built for the computation of the lateral loss (\ie lateral with respect to the car going along the road), which makes the model robust to multi-view datasets like Pitts30k. On the right is shown the construction for the frontal loss, which improves results on frontal-view datasets like MSLS.
    The lateral loss places the \emph{focal point} on the second principal component, whereas the frontal loss places it on the first principal component.
    }
    \label{fig:frontal_lateral_loss}
\end{figure}

Since each cell has two different classes (one \textit{lateral} and one \textit{frontal}), as shown in \cref{fig:frontal_lateral_loss}, we employ two classifiers, one devoted to recognizing viewpoint shifts (with weights $W_{lat}$), and another one tasked with learning frontal-facing (\wrt the vehicle) views (with weights $W_{front}$).
Thus our final loss comprises a \textit{lateral} and a \textit{frontal} component, each relying on a separate CosFace layer. The $\mathcal{L}_\textit{front}$ has the same formulation as $\mathcal{L}_\textit{lat}$. 
Finally, the final loss is:
\begin{equation}
\mathcal{L} = \mathcal{L}_\textit{lat}(f, W_{lat}) + \mathcal{L}_\textit{front}(f, W_{front})
\end{equation}

\section{Experiments}
\label{sec:experiments}


\subsection{Datasets}

\begin{table*}
\begin{center}
\begin{adjustbox}{width=0.9\linewidth}
\centering
\begin{tabular}{l|cccccccccccccccccccccccccccccccc}
\toprule
\multirow{2}{*}{Dataset Name} & \multirow{2}{*}{AmsterTime} & \multirow{2}{*}{Eynsham} & \multirow{2}{*}{Pitts30k} & \multirow{2}{*}{Pitts250k} & Tokyo & San Francisco & SF-XL & SF-XL \\
& & & & & 24/7 & Landmark & test v1 & test v2 \\
\hline
\# queries & 1231 & 24k & 6.8k & 8.3k & 315 & 598 & 1000 & 598 \\
\# database & 1231 & 24k & 10k & 84k & 76k & 1.04M & 2.8M & 2.8M \\
Orientation & multi-view & multi-view & multi-view & multi-view & multi-view & multi-view & multi-view & multi-view \\
Scenery & urban & urban \& country & urban & urban & urban & urban & mostly urban & mostly urban \\
Domain Shift & long-term & none & none & none & day/night & viewpoint & viewpoint, night & viewpoint \\
\bottomrule
\end{tabular}
\end{adjustbox}
\end{center}
\caption{\textbf{Overview of multi-view datasets.} We can see huge variations in size and types of domain shift across the datasets.}
\label{tab:frontal_datasets}
\end{table*}

\begin{table*}
\begin{center}
\begin{adjustbox}{width=0.8\linewidth}
\centering
\begin{tabular}{l|cccccccccccccccccccccccccccccccc}
\toprule
\multirow{2}{*}{Dataset Name} & MSLS & \multirow{2}{*}{Nordland} & \multirow{2}{*}{St Lucia} & SVOX & SVOX & SVOX & SVOX & SVOX \\
& Val & & & Night & Overcast & Rain & Snow & Sun \\
\hline
\# queries & 740 & 27592 & 1464 & 823 & 872 & 937 & 870 & 854 \\
\# database & 18.9k & 27592 & 1549 & 17k & 17k & 17k & 17k & 17k \\
Orientation & frontal-view & frontal-view & frontal-view & frontal-view & frontal-view & frontal-view & frontal-view & frontal-view \\
Scenery & mostly urban & country & suburb & urban & urban & urban & urban & urban \\
Domain Shift & day/night & summer/winter & none & day/night & weather & weather & weather & weather \\
\bottomrule
\end{tabular}
\end{adjustbox}
\end{center}
\caption{\textbf{Overview of frontal-view datasets.} We can see huge variations in size and types of domain shift across the datasets.}
\label{tab:multiview_datasets}
\end{table*}

\begin{figure}
    \begin{center}
    \begin{minipage}{.15\textwidth}
        \includegraphics[width=\textwidth]{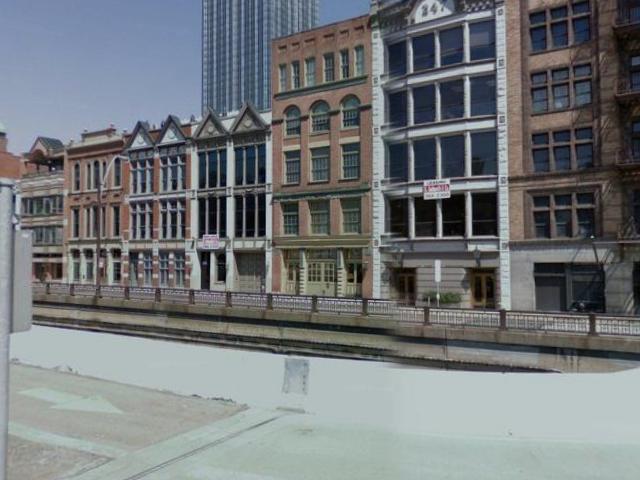}
    \end{minipage}
    \begin{minipage}{.15\textwidth}
        \includegraphics[width=\textwidth]{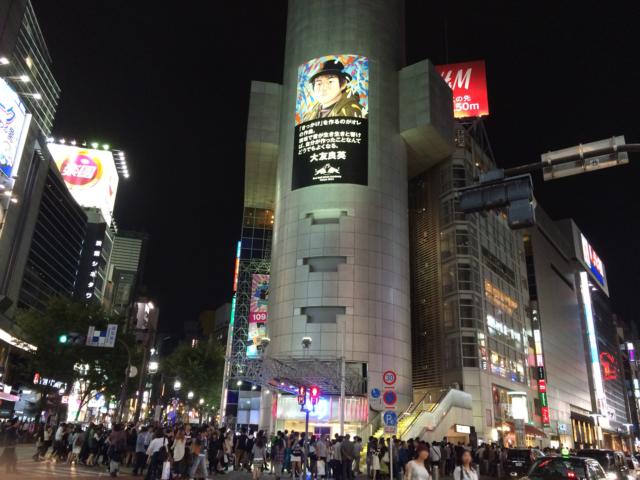}
    \end{minipage}
    \begin{minipage}{.15\textwidth}
        \includegraphics[width=\textwidth]{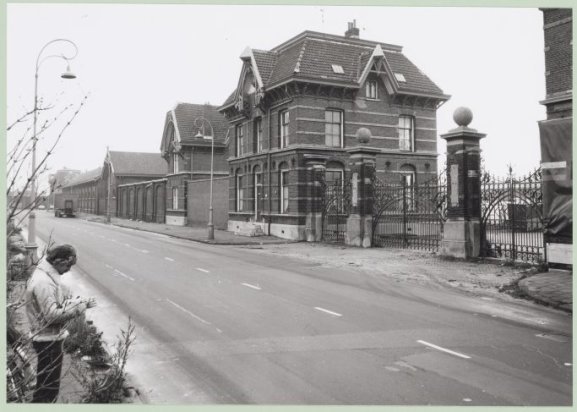}
    \end{minipage}
    \begin{minipage}{.15\textwidth}
        \includegraphics[width=\textwidth]{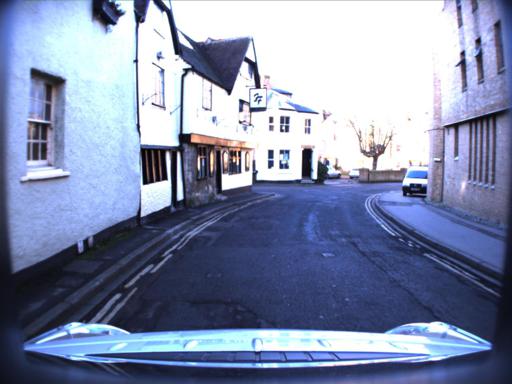}
    \end{minipage}
    \begin{minipage}{.15\textwidth}
    \includegraphics[width=\textwidth]{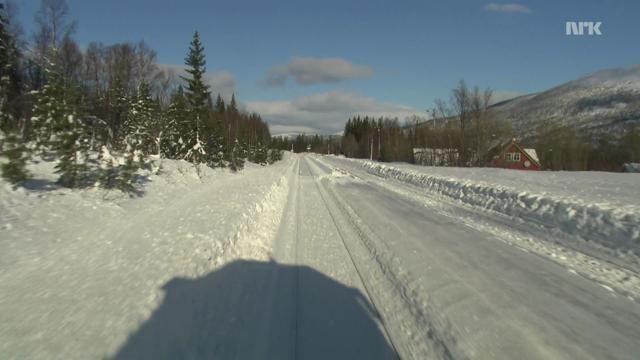}
    \end{minipage}
    \begin{minipage}{.15\textwidth}
    \includegraphics[width=\textwidth]{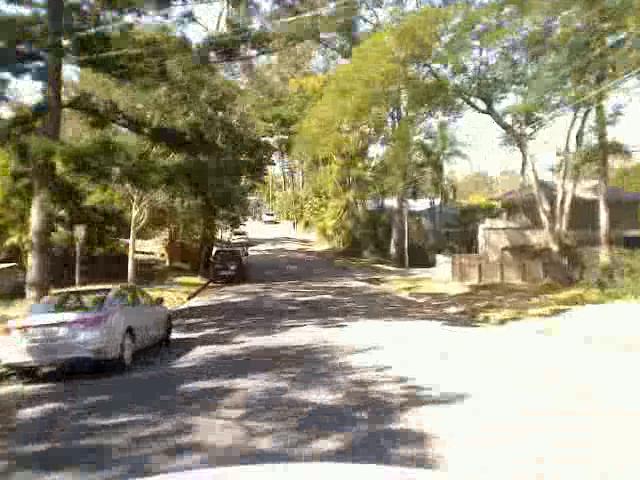}
    \end{minipage}
    \end{center}
    \caption{\textbf{Examples of images from multiple datasets.} In the top row there are queries from multi-view datasets, namely Pitts30k, Tokyo 24/7 and AmsterTime; on the bottom row queries from frontal-view datasets - SVOX sun, Nordland and St Lucia. Further examples from all datasets are reported in the Supplementary.}
    \label{fig:examples}
\end{figure}

To deeply understand the strength and weaknesses of different methods we run experiments on a large number (16) of datasets which present a wide variety of conditions, with various degrees of intra-dataset variability.

Given the large number of datasets, we split them into two categories:
\begin{enumerate}[noitemsep,topsep=1pt]
    \item \textbf{multi-view datasets}, which contain images in any direction \wrt to the direction of the road;
    \item \textbf{frontal-view datasets}, containing for the vast majority images along the road.
\end{enumerate}
The visual differences between the two categories can be seen in \cref{fig:examples}, while the list of datasets is provided in \cref{tab:multiview_datasets} and \cref{tab:frontal_datasets}.

Among the ones with largest \textbf{viewpoint variance}, we note Tokyo 24/7 \cite{Torii_2018_tokyo247}, San Francisco Landmark \cite{Chen_2011_san_francisco} and SF-XL test v1 and v2 \cite{Berton_2022_cosPlace}, all of which contain queries collected with a phone, usually from sidewalks, while the database is from streetview images.
Given the nature of the task, we use mostly urban datasets, with the main exception being Nordland \cite{Sunderhauf_2013_nordland}, which is a collection of photos taken across different seasons with a camera mounted on a train.
Some datasets present various degrees of \textbf{day-to-night} changes, namely MSLS \cite{Warburg_2020_msls}, Tokyo 24/7 \cite{Torii_2018_tokyo247}, SF-XL test v1 \cite{Berton_2022_cosPlace} and SVOX Night \cite{Berton_2021_svox}.
Additionally, SVOX contains a comprehensive set of \textbf{weather domain shifts}, with overcast, rainy, snowy and sunny images.
Eynsham \cite{Cummins_2009_eynsham} is the only completely grayscale dataset, whereas AmsterTime \cite{Yildiz_2022_AmsterTime} contains \textbf{grayscale historical} queries and modern-time RGB database images, making it the only dataset with time variations up to multiple decades.

The datasets are also representative of different sizes of covered area, with the biggest ones being San Francisco Landmark \cite{Chen_2011_san_francisco} (with a database covering $13.6 ~km^{2}$) and SF-XL, which covers $170 ~km^2$.
An overview over each dataset individually is available in the Supplementary.


\subsection{Implementation details}

\myparagraph{Architecture} \\
In order to assess the potential of the EigenPlaces training method in improving the robustness of neural networks, we opt for a very simple architecture made of a standard convolutional neural network (VGG-16 \cite{Simonyan_2015_vgg} or ResNet-50 \cite{He_2016_resnet}, following previous work \cite{Arandjelovic_2018_netvlad, Kim_2017_crn, Liu_2019_sare, Ge_2020_sfrs, Berton_2022_cosPlace, Alibey_2022_gsvcities, Alibey_2023_mixvpr}) to produce embeddings which are fed to a GeM \cite{Radenovic_2019_gem} pooling. Finally a fully connected layer produces the descriptors.
Hence the dimensionality of each descriptors is  equivalent to the number of neurons within the fully connected layer, making it straightforward to change.

This is a much simpler architecture than most previous works, most of which \cite{Arandjelovic_2018_netvlad,Liu_2019_sare,Kim_2017_crn,Ge_2020_sfrs, Peng_2021_appsvr} rely on a more complex NetVLAD layer,
whereas the most recent work, namely MixVPR \cite{Alibey_2023_mixvpr}, employs a MLP-Mixer to aggregate the features provided by the backbone.

\myparagraph{Training} \\
EigenPlaces is trained for 200k iterations with batches of 128 images (64 for each component of the loss).
We use a learning rate of $1e^{-5}$, and as optimizer we use Adam \cite{Kingma_2014_adam}.
We use the same data augmentations as SFRS \cite{Ge_2020_sfrs} (color jittering and random cropping).
Regarding the partitioning in classes, we set M (the side of the squared cells) to 15 meters and $N=3$.
We set the \textit{focal distance} $D=10$ meters following preliminary experiments on the validation set, although in \cref{sec:ablations} we see that using $D=20$ achieves even higher results on average on a number of datasets.

Training is performed on the San Francisco eXtra Large dataset (SF-XL) \cite{Berton_2022_cosPlace}:
to select images pointing towards the \emph{focal point} we first select the whole 360° panorama, from which we obtain a crop with the required orientation.

\myparagraph{Evaluation} \\
Following previous literature \cite{Arandjelovic_2018_netvlad, Kim_2017_crn, Alibey_2022_gsvcities, Alibey_2023_mixvpr, Berton_2022_cosPlace, Masone_2021_survey, Ibrahimi_2021_insideout_vpr, Zaffar_2021_vprbench, Wang_2022_TransVPR,Zhang_2021_gated_netvlad}, we use the recall@N as metric, defined as the percentage of queries for which at least one of the first N predictions is within a given threshold distance.
The threshold is usually set to 25 meters, except for Nordland and AmsterTime.
For Nordland, being the dataset a collection of aligned frames across 4 seasons, a query is considered correctly localized if at least one of its first N predictions is within 10 frames from the ground truth equivalent in the database (as in \cite{Hausler_2019, Hausler_2021_patch_netvlad}).
On the other hand AmsterTime \cite{Yildiz_2022_AmsterTime} is a collection of pairs of images, making a query correctly localized if one of its first N predictions is the query's counterpart in the database.


\subsection{Comparison with previous work}

\begin{table*}
\begin{center}
\begin{adjustbox}{width=0.9\linewidth}
\centering
\begin{tabular}{lccccccccccccccccccc}
\toprule
\multirow{2}{*}{Method} & \multirow{2}{*}{Backbone} & Desc. & \multirow{2}{*}{AmsterTime} & \multirow{2}{*}{Eynsham} & \multirow{2}{*}{Pitts30k} &
\multirow{2}{*}{Pitts250k} & Tokyo & San Francisco & SF-XL & SF-XL \\
& & Dim. & & & & & 24/7 & Landmark & test v1 & test v2 \\
\hline

CosPlace \cite{Berton_2022_cosPlace}     & VGG-16    &  512 &\underline{38.7}&           88.3 &           88.4 &           89.7 &           81.9 &           80.8 &           65.9 &           83.1 \\
\textbf{EigenPlaces (Ours)}              & VGG-16    &  512 &           38.0 &\underline{89.4}&\underline{89.7}&\underline{91.2}&\underline{82.2}&\underline{83.8}&\underline{69.4}&\underline{86.3}\\
\hline
NetVLAD \cite{Arandjelovic_2018_netvlad} & VGG-16    & 4096 &           16.3 &\underline{77.7}&           85.0 &           85.9 &           69.8 &           79.1 &           40.0 &           76.9 \\
SFRS \cite{Ge_2020_sfrs}                 & VGG-16    & 4096 &\underline{29.7}&           72.3 &\underline{89.1}&\underline{90.4}&\underline{80.3}&\underline{83.1}&\underline{50.3}&\underline{83.8}\\
\hline
CosPlace \cite{Berton_2022_cosPlace}     & ResNet-50 &  128 &\underline{39.9}&           88.6 &           89.0 &           89.6 &\underline{81.0}&           82.9 &           69.1 &           86.5 \\
MixVPR \cite{Alibey_2023_mixvpr}         & ResNet-50 &  128 &           23.1 &           84.8 &           87.7 &           88.7 &           56.8 &           66.9 &           36.7 &           68.4 \\
\textbf{EigenPlaces (Ours)}              & ResNet-50 &  128 &           37.9 &\underline{89.1}&\underline{89.6}&\underline{90.2}&           79.4 &\underline{85.5}&\underline{72.4}&\underline{86.6}\\
\hline
CosPlace \cite{Berton_2022_cosPlace}     & ResNet-50 &  512 &\underline{46.4}&           89.9 &           90.2 &           91.7 &           89.5 &           85.6 &           76.7 &           89.0 \\
Conv-AP \cite{Alibey_2022_gsvcities}     & ResNet-50 &  512 &           28.4 &           86.2 &           89.1 &           90.4 &           61.3 &           68.4 &           41.8 &           64.0 \\
MixVPR \cite{Alibey_2023_mixvpr}         & ResNet-50 &  512 &           35.8 &           87.6 &           90.4 &           93.0 &           78.4 &           79.4 &           57.7 &           84.3 \\
\textbf{EigenPlaces (Ours)}              & ResNet-50 &  512 &           45.7 &\underline{90.5}&\underline{91.9}&\underline{93.5}&\underline{89.8}&\underline{89.5}&\underline{82.6}&\underline{90.6}\\
\hline
CosPlace \cite{Berton_2022_cosPlace}     & ResNet-50 & 2048 &           47.7 &           90.0 &           90.9 &           92.3 &           87.3 &           87.1 &           76.4 &           88.8 \\
Conv-AP \cite{Alibey_2022_gsvcities}     & ResNet-50 & 2048 &           31.3 &           86.6 &           90.4 &           92.3 &           71.1 &           71.7 &           47.8 &           68.1 \\
\textbf{EigenPlaces (Ours)}              & ResNet-50 & 2048 &  \textbf{48.9} &  \textbf{90.7} &  \textbf{92.5} &  \textbf{94.1} &  \textbf{93.0} &  \textbf{89.6} &  \textbf{84.1} &  \textbf{90.8} \\
\hline
Conv-AP \cite{Alibey_2022_gsvcities}     & ResNet-50 & 4096 &           33.9 &           87.5 &           90.5 &           92.3 &           76.2 &           73.7 &           47.5 &           74.4 \\
MixVPR \cite{Alibey_2023_mixvpr}         & ResNet-50 & 4096 &\underline{40.2}&\underline{89.4}&\underline{91.5}&  \textbf{94.1} &\underline{85.1}&\underline{83.8}&\underline{71.1}&\underline{88.5}\\
Conv-AP \cite{Alibey_2022_gsvcities}     & ResNet-50 & 8192 &           35.0 &           87.6 &           90.5 &           92.6 &           72.1 &           74.4 &           49.3 &           75.8 \\

\bottomrule
\end{tabular}
\end{adjustbox}
\end{center}
\caption{\textbf{Recall@1 on multi-view datasets}, split according to the utilized backbone and descriptors dimension. Best overall results on each dataset are in \textbf{bold}, best results for each group are \underline{underlined}.}
\label{tab:multi_view}
\end{table*}

\begin{table*}
\begin{center}
\begin{adjustbox}{width=0.8\linewidth}
\centering
\begin{tabular}{lccccccccccccccccccc}
\toprule
\multirow{2}{*}{Method} & \multirow{2}{*}{Backbone} & Desc. & MSLS &
\multirow{2}{*}{Nordland} & \multirow{2}{*}{St Lucia} & SVOX & SVOX & SVOX & SVOX & SVOX \\
& & Dim. & Val & & & Night & Overcast & Rain & Snow & Sun \\
\hline

CosPlace \cite{Berton_2022_cosPlace}     & VGG-16    &  512 &           82.6 &\underline{58.5}&           95.3 &\underline{44.8}&           88.5 &\underline{85.2}&           89.0 &           67.3 \\
\textbf{EigenPlaces (Ours)}              & VGG-16    &  512 &\underline{84.2}&           54.5 &\underline{95.4}&           42.3 &\underline{89.4}&           83.5 &\underline{89.2}&\underline{69.7}\\
\hline
NetVLAD \cite{Arandjelovic_2018_netvlad} & VGG-16    & 4096 &           58.9 &           13.1 &           64.6 &            8.0 &           66.4 &           51.5 &           54.4 &           35.4 \\
SFRS \cite{Ge_2020_sfrs}                 & VGG-16    & 4096 &\underline{70.0}&\underline{16.0}&\underline{75.9}&\underline{28.6}&\underline{81.1}&\underline{69.7}&\underline{76.0}&\underline{54.8}\\
\hline
CosPlace \cite{Berton_2022_cosPlace}     & ResNet-50 &  128 &\underline{85.5}&\underline{54.7}&           98.7 &\underline{35.4}&           88.5 &           80.4 &           86.6 &           65.2 \\
MixVPR \cite{Alibey_2023_mixvpr}         & ResNet-50 &  128 &           79.1 &           47.8 &\underline{99.0}&           25.9 &\underline{92.3}&           80.9 &           87.7 &\underline{73.5}\\
\textbf{EigenPlaces (Ours)}              & ResNet-50 &  128 &           83.4 &           50.5 &           98.8 &           29.0 &           90.9 &\underline{83.8}&\underline{91.1}&           68.5 \\
\hline
CosPlace \cite{Berton_2022_cosPlace}     & ResNet-50 &  512 &           86.9 &           66.5 &           99.1 &\underline{51.6}&           90.0 &           87.3 &           89.5 &           75.9 \\
Conv-AP \cite{Alibey_2022_gsvcities}     & ResNet-50 &  512 &           82.3 &           59.2 &           99.2 &           36.0 &           90.5 &           80.3 &           86.4 &           75.3 \\
MixVPR \cite{Alibey_2023_mixvpr}         & ResNet-50 &  512 &           83.6 &           67.2 &           99.2 &           44.8 &\underline{93.9}&           86.4 &\underline{93.9}&           78.7 \\
\textbf{EigenPlaces (Ours)}              & ResNet-50 &  512 &  \textbf{89.5} &\underline{67.9}&\underline{99.5}&           51.5 &           92.8 &\underline{89.0}&           92.0 &\underline{83.1}\\
\hline
CosPlace \cite{Berton_2022_cosPlace}     & ResNet-50 & 2048 &           87.4 &\underline{71.9}&\underline{99.6}&           50.7 &           92.2 &           87.0 &           92.0 &           78.5 \\
Conv-AP \cite{Alibey_2022_gsvcities}     & ResNet-50 & 2048 &           81.2 &           62.3 &           99.3 &           37.9 &           92.0 &           83.7 &           90.2 &           80.3 \\
\textbf{EigenPlaces (Ours)}              & ResNet-50 & 2048 &\underline{89.1}&           71.2 &\underline{99.6}&\underline{58.9}&\underline{93.1}&\underline{90.0}&\underline{93.1}&  \textbf{86.4} \\
\hline
Conv-AP \cite{Alibey_2022_gsvcities}     & ResNet-50 & 4096 &           82.8 &           59.6 &           99.6 &           41.9 &           91.2 &           81.9 &           87.9 &           82.0 \\
MixVPR \cite{Alibey_2023_mixvpr}         & ResNet-50 & 4096 &\underline{87.2}&  \textbf{76.2} &           99.6 &  \textbf{64.4} &  \textbf{96.2} &  \textbf{91.5} &  \textbf{96.8} &\underline{84.8}\\
Conv-AP \cite{Alibey_2022_gsvcities}     & ResNet-50 & 8192 &           82.4 &           62.9 &  \textbf{99.7} &           43.4 &           91.9 &           82.8 &           91.0 &           80.4 \\

\bottomrule
\end{tabular}
\end{adjustbox}
\end{center}
\caption{\textbf{Recall@1 on frontal-view datasets}, split according to the utilized backbone and descriptors dimension. Best overall results on each dataset are in \textbf{bold}, best results for each group are \underline{underlined}.}
\label{tab:frontal_view}
\end{table*}

\myparagraph{Baselines.}
We run an extensive set of experiments to thoroughly evaluate the soundness of EigenPlaces, comparing it with a large number of open source methods from the literature.
Specifically, we use older NetVLAD-based methods which rely on a VGG-16 backbone, namely NetVLAD itself \cite{Arandjelovic_2018_netvlad} and SFRS \cite{Ge_2020_sfrs}.
We also compute results with the more recent CosPlace \cite{Berton_2022_cosPlace}, and the latest works of Conv-AP \cite{Alibey_2022_gsvcities} and MixVPR \cite{Alibey_2023_mixvpr}, which were trained on the Google StreetView (GSV) dataset \cite{Alibey_2022_gsvcities}.
CosPlace, Conv-AP and MixVPR provide open-source models with multiple backbone and descriptors dimensionalities, allowing us to provide a large number of comparisons with different architectures.

Following previous VPR works that use image retrieval \cite{Arandjelovic_2018_netvlad, Kim_2017_crn, Liu_2019_sare, Zhu_2018_apanet, Ge_2020_sfrs, Berton_2022_cosPlace, Alibey_2022_gsvcities, Peng_2021_appsvr, Peng_2021_sralNet, Yu_2020_SPEVlad}, we do not compare pure retrieval methods like the ones in \cref{tab:multi_view} with 2-stage re-ranking techniques, such as \cite{Hausler_2021_patch_netvlad, Wang_2022_TransVPR, Cao_2020_delg, Fuwen_2021_reranking_transformers}.

\myparagraph{Discussion of results.}
Given the large number of datasets, we split the results in two parts:
\begin{enumerate}[noitemsep,topsep=1pt]
    \item in \cref{tab:multi_view} we show results on multi-view datasets, where the database and queries orientation can vary across $360 \degree$ ;
    \item in \cref{tab:frontal_view} we report experiments on frontal-view datasets, \ie where the vast majority (or all) of the images are forward facing.
\end{enumerate}
The findings from our large set of experiments can be summarized as follows:
\begin{itemize}[noitemsep,topsep=1pt]
    \item Firstly we can see that older methods like NetVLAD (2016) and SFRS (2020), despite producing larger descriptors are less robust to domain shifts, and are easily outperformed by newer models especially on frontal-view datasets.
    \item Latest models, namely CosPlace (2022), Conv-AP (2022) and MixVPR (2023), all provide robust results even when using compact descriptors.
    \item There is no single model that outperforms all other ones on all datasets, and different models have different characteristics and strenghts.
    \item Despite not achieving state of the art on all datasets, EigenPlaces has the best overall results, which is especially noticeable on multi-view datasets from \cref{tab:multi_view} which provide larger viewpoint changes.
    \item MixVPR on average outperforms EigenPlaces on frontal-view datasets, although at the cost of twice bigger descriptors.
    \item EigenPlaces and CosPlace provide strong results also on datasets with grayscale images, namely AmsterTime and Eynsham, despite not being trained with grayscale augmentation.
    \item Among the most interesting findings, we can see that with low-dimensionality descriptors (\ie 128-D) CosPlace, MixVPR and EigenPlaces provide remarkably good results on datasets with little to no domain shift between database and queries (\eg Pitts30k, St Lucia), although lower dimensionality descriptors still struggle on cross-domain datasets (\eg AmsterTime, Tokyo 24/7, SVOX night).
    \item With the impressive results reached in the last two years, many datasets can be considered almost solved, with results reaching over 90\% in Recall@1, with Recall@10 higher than 95\% in most cases (see the Supplementary).
\end{itemize}
A more extensive set of experiments is reported in the Supplementary.


\subsection{Analysis of resources}

\myparagraph{GPU memory footprint.}
EigenPlaces is surprisingly cheap to train, as it can train its best architecture using less than 7 GB of memory (ResNet-50 with 2048-D descriptors).
This makes it quite lighter than MixVPR \cite{Alibey_2023_mixvpr}, which requires more than 18 GB of memory, using their batch size of 480 (\ie 120 quadruplets).
Following previous work \cite{Alibey_2022_gsvcities, Alibey_2023_mixvpr} we use mixed precision to reduce GPU footprint and speed up computation.

\myparagraph{Training and evaluation time.}
EigenPlaces with our best architecture takes 24 hours to train on a single 3090 GPU, which is similar to the duration of training previous methods (SFRS, CosPlace, MixVPR), and we found that different descriptors dimensionality have a negligible impact on training time.

On the other hand, descriptors dimensionality is linearly correlated to two very important factors in large scale image retrieval: \textbf{memory footprint} and \textbf{matching time} (\ie the time required by the nearest neighbor search).
Therefore, wrt MixVPR's best configuration, our top performing model is twice as fast and requires half the memory, while achieving overall better results.
Note that in real-world systems the descriptors of the database images are extracted offline, and the inference time can be computed as the sum of the query's descriptors extraction time plus the matching (kNN) time, rendering the extraction time negligible when working on large scale datasets.


\subsection{Ablations}
\label{sec:ablations}

\begin{table}
\begin{center}
\begin{adjustbox}{width=0.95\linewidth}
\begin{tabular}{ccccccc}
\toprule
\multirow{2}{*}{\begin{tabular}[c]{@{}c@{}}Lateral\\Loss\end{tabular}} &
\multirow{2}{*}{\begin{tabular}[c]{@{}c@{}}Frontal\\Loss\end{tabular}} &
\multicolumn{1}{c}{\multirow{2}{*}{Pitts30k}} &
\multicolumn{1}{c}{\multirow{2}{*}{Tokyo 24/7}} &
\multicolumn{1}{c}{\multirow{2}{*}{MSLS Val}} &
\multicolumn{1}{c}{\multirow{2}{*}{St Lucia}} &
\multicolumn{1}{c}{\multirow{2}{*}{Average}} \\ \\
\hline
\checkmark &            & \underline{90.2} & \underline{80.0} & 83.1 & 97.3 & 87.6 \\
  & \checkmark          & 89.5 & 78.1 & \underline{85.8} & \textbf{99.3} & 88.2 \\
\checkmark & \checkmark & \textbf{90.5} & \textbf{82.2} & \textbf{86.2} & \underline{99.0} & \textbf{89.5} \\
\bottomrule
\end{tabular}
\end{adjustbox}
\end{center}
\caption{\textbf{Ablation on the two components of the loss.} Experiments show the Recall@1 obtained with a ResNet-18 with output dimensionality 512. We can see that training with the frontal loss only achieves good results on images that are mostly made of frontal-view images (MSLS and St Lucia) but poor  on others, and the model with both components of the loss achieves best overall performances.}
\label{tab:ablation_losses}
\end{table}

\myparagraph{Ablation on the loss.}
In this section we investigate how each of the two components of the loss affects the results.
An ablation is reported in \cref{tab:ablation_losses}.
Experimental evidence shows that using only the lateral loss, which places the \textit{focal point} on the second principal component (see \cref{fig:frontal_lateral_loss}), is enough to reach satisfactory results on multi-view datasets like Pitts30k and Tokyo 24/7, although it fails to produce robust embeddings for frontal-view datasets.
On the other hand, relying solely on the frontal-view loss, which places the \textit{focal point} on the first principal component, allows to attain very strong results on MSLS and St Lucia. On Pitts30k and Tokyo 24/7, this configuration suffers from a considerable drop.
Finally, their combination provides a robust combination of each component's strength, and reaches good results on all datasets.

\myparagraph{Ablation on the focal distance.}
In this section we compute experiments changing the \textit{focal distance}, \ie the distance between the mean of the images' position and the \textit{focal point}. Results are in \cref{tab:focal_distances}.
Although the best overall results are achieved with a \textit{focal distance} of 20 and 10 meters, using higher distances leads to good results on frontal-view datasets.
This is not surprising, given that higher focal distances lead the training images' orientation to be further from the center of the cells (\ie straight along the road), as is usually the case for these kind of datasets.
On the other hand, we can see that a \textit{focal distance} of 0 meters achieves better results than expected, considering that in this situation some of the images will be facing opposite directions.

\begin{table}
\begin{center}
\begin{adjustbox}{width=0.9\linewidth}
\begin{tabular}{cccccc}
\toprule
\multirow{2}{*}{\begin{tabular}[c]{@{}c@{}}Focal Distance\\(meters)\end{tabular}} &
\multicolumn{1}{c}{\multirow{2}{*}{Pitts30k}} &
\multicolumn{1}{c}{\multirow{2}{*}{Tokyo 24/7}} &
\multicolumn{1}{c}{\multirow{2}{*}{MSLS Val}} &
\multicolumn{1}{c}{\multirow{2}{*}{St Lucia}} & 
\multicolumn{1}{c}{\multirow{2}{*}{Average}} \\ \\
\hline
 0 & 89.4 & 74.0 & 82.6 & 98.4 & 86.1 \\
10 & \textbf{90.5} & 82.2 & \textbf{86.2} & 99.0 & 89.5 \\
20 & 90.3 & \textbf{84.4} & 86.1 & \textbf{99.5} & \textbf{90.1} \\
30 & 90.3 & 82.9 & 85.0 & \textbf{99.5} & 89.4 \\
50 & 90.4 & 83.8 & 85.9 & \textbf{99.5} & 89.9 \\
\bottomrule
\end{tabular}
\end{adjustbox}
\end{center}
\caption{\textbf{Ablation on focal distance}, shown as the Recall@1 obtained with a ResNet-18 with output dimensionality 512 on multiple datasets.}
\label{tab:focal_distances}
\end{table}

\begin{figure}
    \begin{center}
    \includegraphics[width=0.48\textwidth]{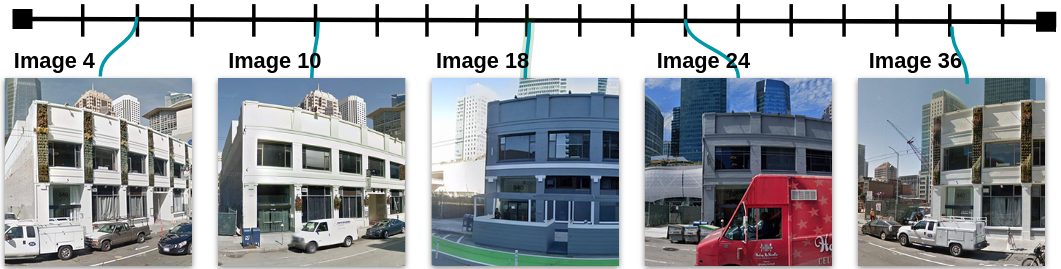}
    \includegraphics[width=0.48\textwidth]{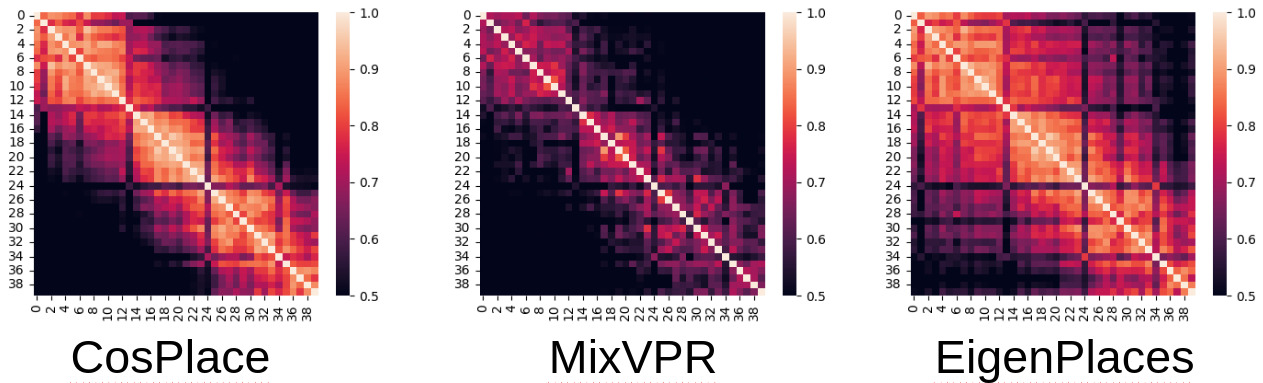}
    \end{center}
    \caption{
    \textbf{Confusion matrices with the cosine similarity among 40 images representing the same place, from different viewpoints.} The cosine similarity is computed in features space, with the three most relevant methods of CosPlace, MixVPR and EigenPlaces.
    For example, the value withing the matrix at position (2, 34) is the cosine similarity between the second and 34th image within a given cell.    
    We can see that EigenPlaces is able to have high correlation even from images that have very different viewpoint (\ie with indexes distant from each other), whereas previous works only share similar features among images that have very close point of view. Some descriptors are quite different from the others as those images might have occlusion (\eg the red truck in image 24).}
    \label{fig:confusion_matrices}
\end{figure}

\myparagraph{Embeddings invariance.}
In \cref{fig:confusion_matrices} we test whether our proposed training algorithm is indeed effective in embedding viewpoint robustness in the model. In a randomly selected cell, we sort the images along the first principal component, extract the features of the images oriented towards the \textit{focal point} and compute a similarity matrix. The obtained matrix shows along its rows and columns what happens to the embeddings when traversing the principal component. For previous works, this analysis shows clearly a rapid decrease in embedding similarity when changing the viewpoint, whereas EigenPlaces ensures more robustness.


\section{Conclusions}
\label{sec:conclusions}

In this work we introduced a novel training algorithm for VPR, that tackles the challenge of perspective shifts.
After dividing the available map into fine-grained cells, our method builds classes by inferring from the data inside each cell a point of interest that is depicted from as many different viewpoints as possible. By minimizing a loss that asks the network to recognize the same point from various perspectives, we embed viewpoint-invariance into a feature extractor.
We support our contribution through extensive experiments on a vast amount of datasets with diverse characteristics and challenges. We discuss how each dataset can highlight different capabilities in a model, and despite the wide variety of test cases we show that using EigenPlaces we obtain SOTA result in the majority of cases, while using lighter descriptors than previous works.

{\small
\bibliographystyle{ieee_fullname}
\bibliography{egbib}
}

\appendix

\section*{Supplementary}
In this supplementary material we show:
\begin{itemize}[noitemsep,topsep=1pt]
    \item in \cref{sec:supp_different_training_methods} how different training methods use different images at train time;
    \item in \cref{sec:supp_datasets} we provide further information regarding the datasets;
    \item in \cref{sec:supp_experiments} further quantitative and qualitative results from our large set of experiments.
\end{itemize}

\section{Data for Different Training Methods}
\label{sec:supp_different_training_methods}

\begin{figure}
    \begin{center}
    \begin{subfigure}{\linewidth}
        \includegraphics[width=\linewidth]{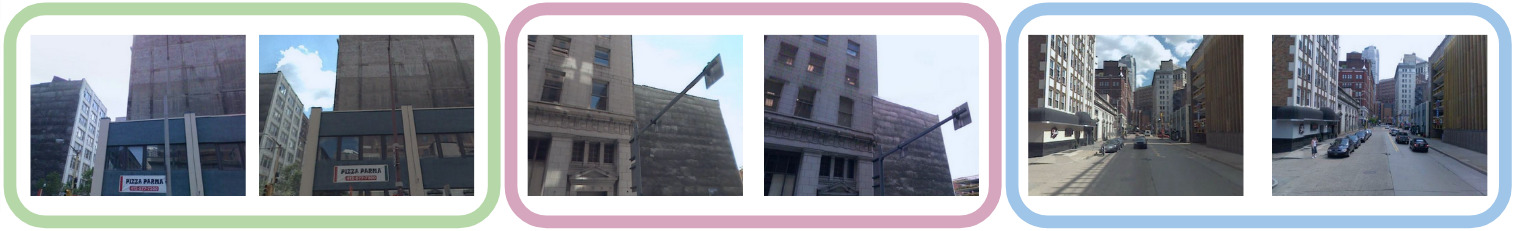}
        \caption{Three Query-positive pairs mined as in \textbf{NetVLAD}}
    \end{subfigure}
    \par\bigskip
    \begin{subfigure}{\linewidth}
        \includegraphics[width=\linewidth]{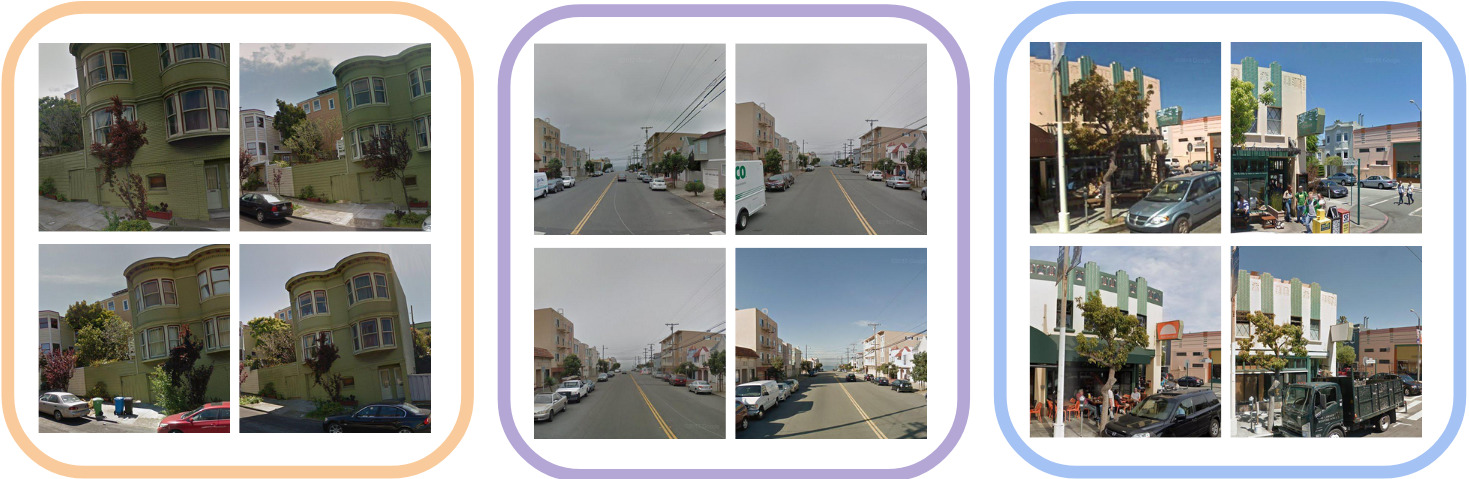}
        \caption{Images within three classes, created with \textbf{CosPlace}}
    \end{subfigure}
    \par\bigskip
    \begin{subfigure}{\linewidth}
        \includegraphics[width=\linewidth]{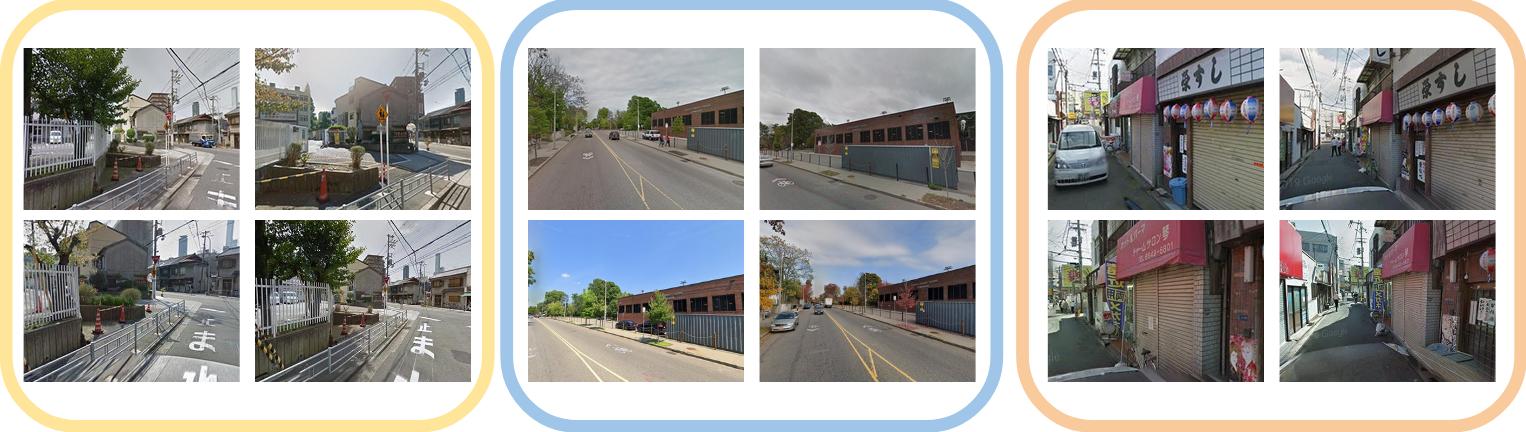}
        \caption{Images within three classes, as used for training of \textbf{Conv-AP} and \textbf{MixVPR}}
    \end{subfigure}
    \par\bigskip
    \begin{subfigure}{\linewidth}
        \includegraphics[width=\linewidth]{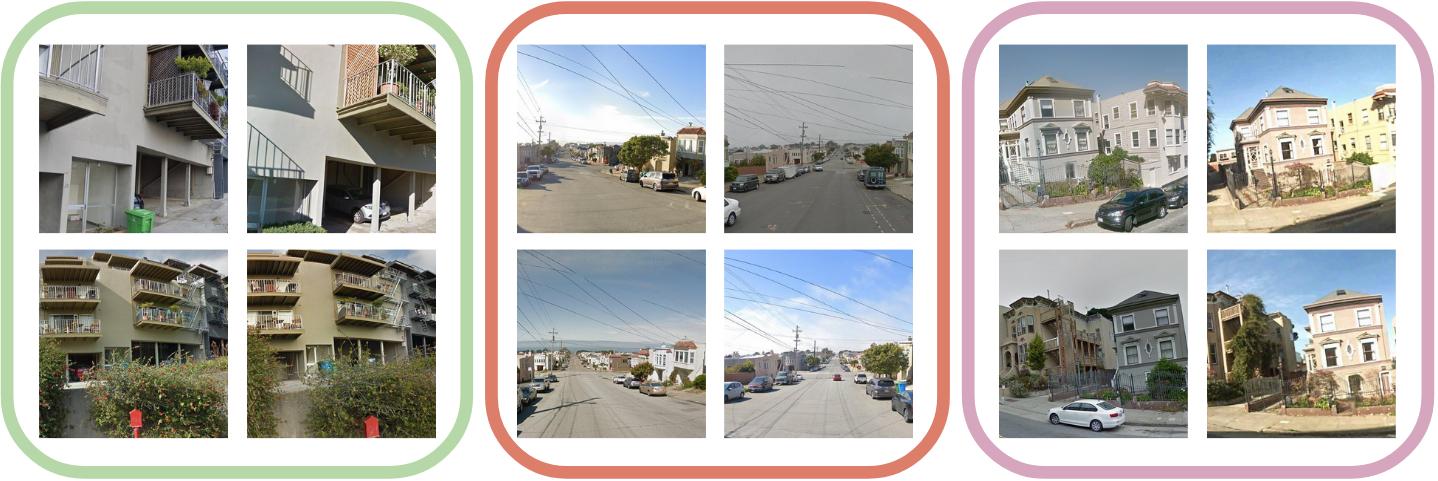}
        \caption{Training data from three classes created with \textbf{EigenPlaces}}
    \end{subfigure}
    \end{center}
    \caption{
    \textbf{Training data with different methods.} Only the images used by EigenPlaces provide large viewpoint shifts.
    }
    \label{fig:supp_training_examples}
\end{figure}

Visualization of the training data used by different training methods is shown in \cref{fig:supp_training_examples}.

In the image we can see that the query-positive pairs mined with NetVLAD \cite{Arandjelovic_2018_netvlad} have very little viewpoint shift. 
NetVLAD \cite{Zhang_2021_gated_netvlad} uses positive mining to obtain the most similar positive to the query, which is then used within a triplet loss. Note that this is different from negative mining (which in NetVLAD is also performed. This is the same (or very similar) approach used by most following works \cite{Kim_2017_crn, Liu_2019_sare, Ge_2020_sfrs, Zhu_2018_apanet, Peng_2021_appsvr, Peng_2021_sralNet, Hausler_2021_patch_netvlad}.

CosPlace \cite{Berton_2022_cosPlace} uses images with the same orientation for a given class, with images being just a few meters apart from each other.
Conv-AP \cite{Alibey_2022_gsvcities} and MixVPR \cite{Alibey_2023_mixvpr} use a pre-defined set of classes by GSV-Cities \cite{Alibey_2022_gsvcities}, with little intra-class viewpoint variations.

In contrast with previous methods, EigenPlaces creates training data by ensuring large viewpoint shifts between images, as visually shown in the last row of \cref{fig:supp_training_examples}, which in turn make the trained model more robust.

\section{Datasets}
\label{sec:supp_datasets}

We test all models on a large number of datasets, which helps to thoroughly understand each method's strength and weaknesses.
To download a number of datasets (namely AmsterTime, Eynsham, San Francisco Landmark, Nordland, St Lucia and SVOX) we used the open-source automatic downloader from \cite{Berton_2022_benchmark}, as this ensures maximum reproducibility for future research.
Below is a short description for each of the datasets.

\myparagraph{AmsterTime} \cite{Yildiz_2022_AmsterTime}
is a collection of over one thousand pairs of query-reference images from the city of Amsterdam.
For each pair, the query is a grayscale historical image, and its reference is a modern-day photo which represents the same place, as confirmed by human experts.
The pairs provide multiple domain shifts: viewpoints, long-term temporal changes, modality (RGB vs grayscale), different cameras.
This makes AmsterTime one of the most challenging dataset available, despite its relatively small scale.
\vspace{-0.3cm}
\begin{figure}[H]
    \begin{center}
    \includegraphics[width=0.4\linewidth]{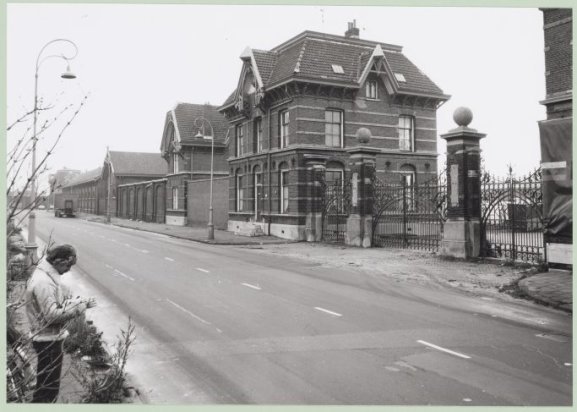}
    \includegraphics[width=0.4\linewidth]{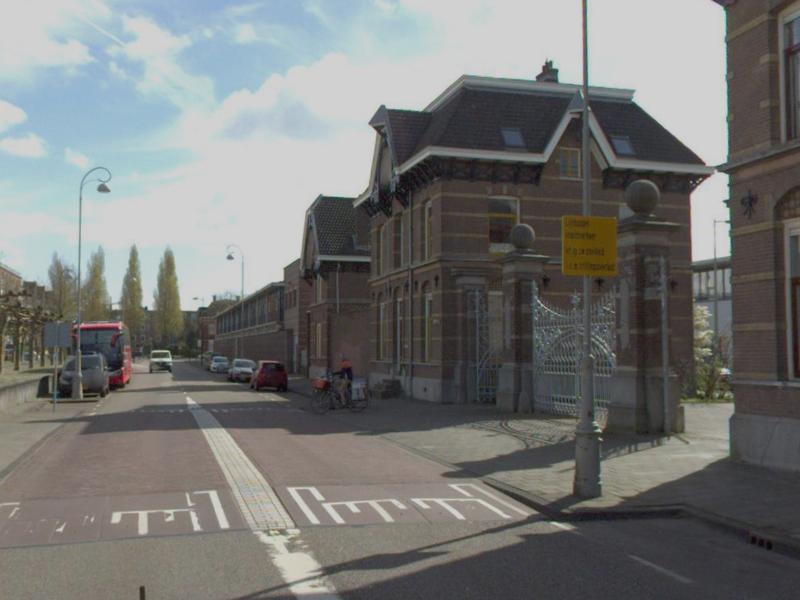}
    \end{center}
    \vspace{-0.6cm}
    \caption{Examples from AmsterTime query and database.}
\end{figure}
\vspace{-0.3cm}

\myparagraph{Eynsham} \cite{Cummins_2009_eynsham}
consists of images from cameras mounted on a car and GPS co-ordinates of the car going around around a loop twice.
The original images are 360° panoramas, that we split in crops following standard practice \cite{Torii_2015_pitts250k, Torii_2018_tokyo247, Berton_2022_benchmark}.
The images are grayscale, and the car drives around the Oxford countryside, passing also through the city of Oxford.
\vspace{-0.3cm}
\begin{figure}[H]
    \begin{center}
    \includegraphics[width=0.4\linewidth]{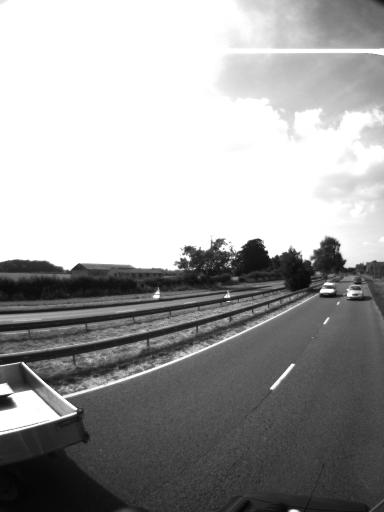}
    \includegraphics[width=0.4\linewidth]{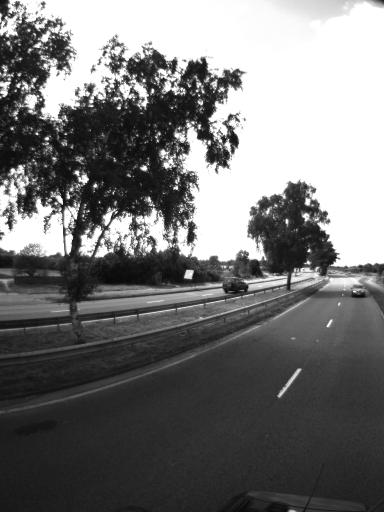}
    \end{center}
    \vspace{-0.6cm}
    \caption{Examples from Eynsham query and database.}
\end{figure}
\vspace{-0.3cm}

\myparagraph{Pitts30k and Pitts250k} \cite{Torii_2015_pitts250k}
are perhaps the most used dataset for VPR to date, on which a large number of works present their results \cite{Arandjelovic_2018_netvlad, Kim_2017_crn, Liu_2019_sare, Ge_2020_sfrs, Alibey_2022_gsvcities, Alibey_2023_mixvpr, Berton_2021_geowarp, Hausler_2021_patch_netvlad, Zhang_2021_gated_netvlad, Berton_2022_cosPlace}.
They are built with Google StreetView images from the city center of Pittsburgh, by ensuring that database and queries are taken in different years.
They provide three splits for training, validation and test.
The 6816 test queries used for Pitts30k are a subset of the 8280 used for Pitts250k, whereas the Pitts250k database is roughly 8 times larger.
\vspace{-0.3cm}
\begin{figure}[H]
    \begin{center}
    \includegraphics[width=0.4\linewidth]{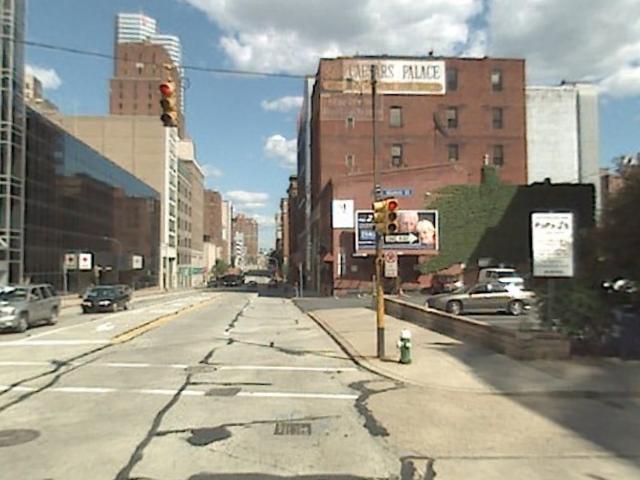}
    \includegraphics[width=0.4\linewidth]{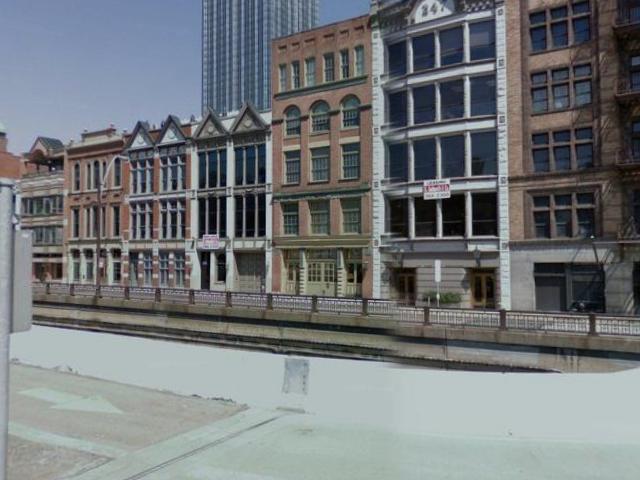}
    \end{center}
    \vspace{-0.6cm}
    \caption{Examples from Pitts30k query and database.}
\end{figure}
\vspace{-0.3cm}

\myparagraph{Tokyo 24/7} \cite{Torii_2018_tokyo247} is a challenging dataset from the center of Tokyo.
The database is made from Google StreetView, whereas the queries are a collection of smartphone photos from 105 places, and each place is photographed during the day, at sunset and at night. This results in 315 queries, each to be geolocalized independently.
\vspace{-0.3cm}
\begin{figure}[H]
    \begin{center}
    \includegraphics[width=0.4\linewidth]{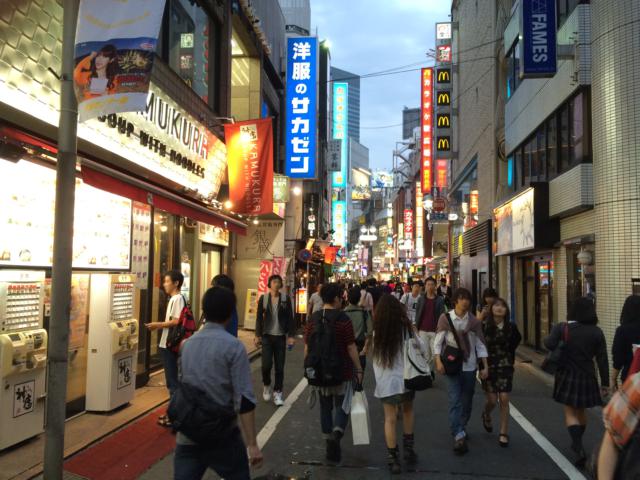}
    \includegraphics[width=0.4\linewidth]{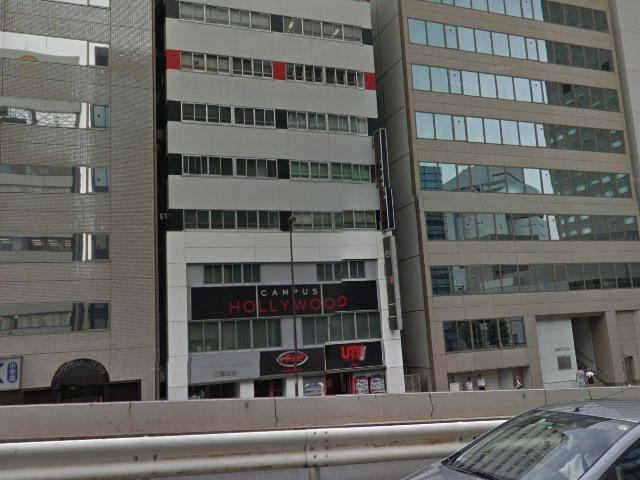}
    \end{center}
    \vspace{-0.6cm}
    \caption{Examples from Tokyo 24/7 query and database.}
\end{figure}
\vspace{-0.3cm}

\myparagraph{San Francisco Landmark} \cite{Chen_2011_san_francisco} is a large dataset from the center of San Francisco with a database of more than 1M images, and a set of 598 queries collected with a smartphone.
\vspace{-0.3cm}
\begin{figure}[H]
    \begin{center}
    \includegraphics[width=0.4\linewidth]{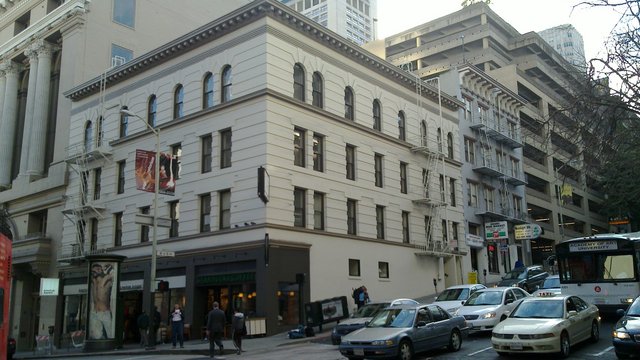}
    \includegraphics[width=0.4\linewidth]{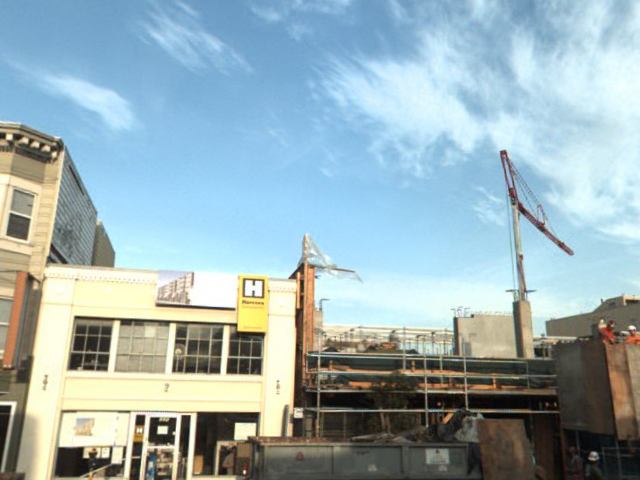}
    \end{center}
    \vspace{-0.6cm}
    \caption{Examples from San Francisco Landmark query and database.}
\end{figure}
\vspace{-0.3cm}

\myparagraph{San Francisco eXtra Large} (SF-XL) \cite{Berton_2022_cosPlace} is a huge dataset covering the whole city of San Francisco with over 41M images. Its test set covers the same with a less dense set of 2.8M images.
Two sets of queries are used: the first (\textit{test v1}) is a challenging set of 1000 images from Flickr, with multiple challenges like night images and photos from the sidewalk. \textit{Test v2} uses the same set of queries from San Francisco Landmark.
\vspace{-0.3cm}
\begin{figure}[H]
    \begin{center}
    \includegraphics[width=0.2\linewidth]{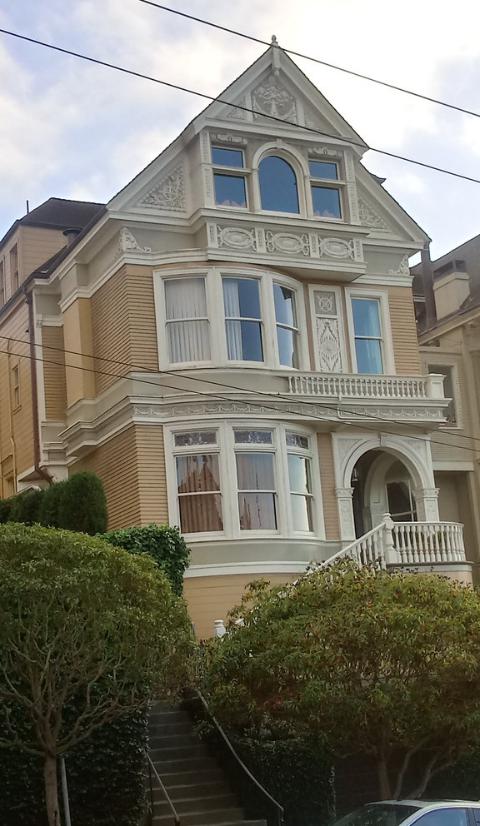}
    \includegraphics[width=0.3\linewidth]{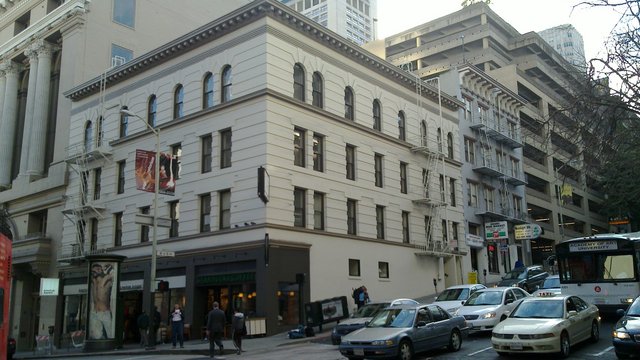}
    \includegraphics[width=0.3\linewidth]{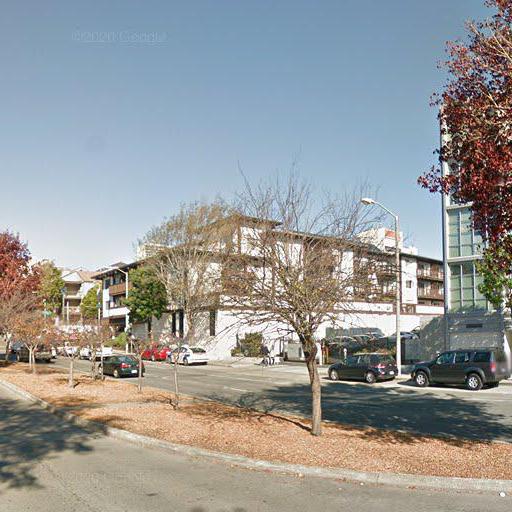}
    \end{center}
    \vspace{-0.6cm}
    \caption{Examples from SF-XL: (left to right) a query from SF-XL \textit{test v1}, a query from SF-XL \textit{test v2} and an example from database.}
\end{figure}
\vspace{-0.3cm}

\myparagraph{MSLS} \cite{Warburg_2020_msls} is the Mapillary Street-Level Sequences dataset, which has been created for image and sequence-based VPR. The dataset consists of more than 1M images from multiple cities, although only a small subset is used for evaluation. Following common practice \cite{Hausler_2021_patch_netvlad, Alibey_2023_mixvpr, Alibey_2022_gsvcities} we evaluate on their validation set, as the labels for the test set have not been released. The test set is from the cities of Copenhagen and San Francisco, and although being mostly single-domain, it provides a small number of night and lateral-view images.
\vspace{-0.3cm}
\begin{figure}[H]
    \begin{center}
    \includegraphics[width=0.4\linewidth]{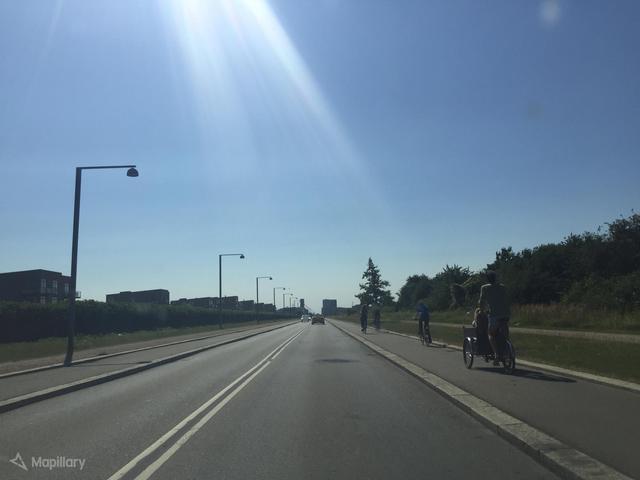}
    \includegraphics[width=0.4\linewidth]{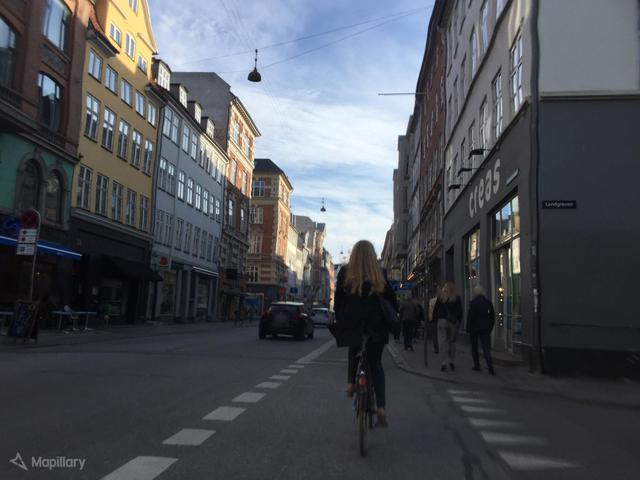}
    \end{center}
    \vspace{-0.6cm}
    \caption{Examples from MSLS query and database.}
\end{figure}
\vspace{-0.3cm}

\myparagraph{Nordland} \cite{Sunderhauf_2013_nordland} was collected by recording a video from a train riding through the Norwegian countryside, and traversing the same path across four seasons.
Images are then extracted at 1FPS.
Following previous works \cite{Hausler_2019, Hausler_2021_patch_netvlad} we use the winter traverse as queries and summer as database, which have been post-processed to ensure alignment of the frames. Unlike in most other VPR datasets, a query is considered correctly localized if the matched database image is less than 10 frames away.
\vspace{-0.3cm}
\begin{figure}[H]
    \begin{center}
    \includegraphics[width=0.4\linewidth]{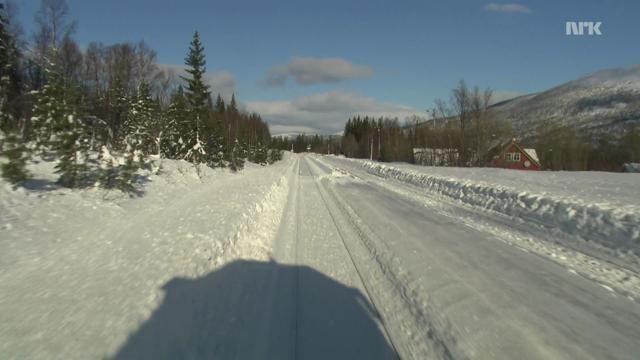}
    \includegraphics[width=0.4\linewidth]{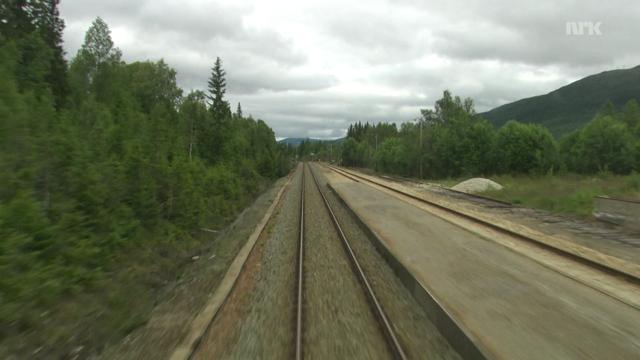}
    \end{center}
    \vspace{-0.6cm}
    \caption{Examples from Nordland query and database.}
\end{figure}
\vspace{-0.3cm}

\myparagraph{St Lucia} \cite{Milford_2008_st_lucia}
is a dataset collected with a car-mounted camera, with long videos from multiple drives along the same area: the St Lucia suburb of Brisbane.
Following \cite{Berton_2022_benchmark}, we use the first and last drive (of the nine available) as queries and database, and we sample one frame every 5 meters of driving.
\vspace{-0.3cm}
\begin{figure}[H]
    \begin{center}
    \includegraphics[width=0.4\linewidth]{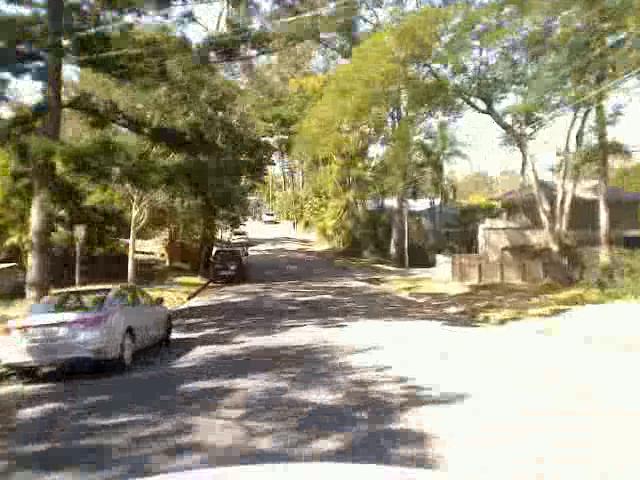}
    \includegraphics[width=0.4\linewidth]{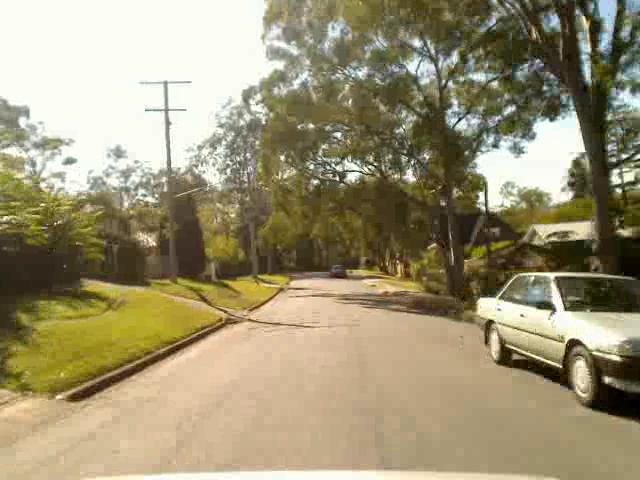}
    \end{center}
    \vspace{-0.6cm}
    \caption{Examples from St Lucia query and database.}
\end{figure}
\vspace{-0.3cm}

\myparagraph{SVOX} \cite{Berton_2021_svox}
is a cross-domain dataset built from cross-domain VPR, that allows to evaluate on multiple weather conditions.
It spans the city of Oxford, with a large (single-domain) database from Google StreetView images: the queries are instead from the Oxford RobotCar dataset \cite{Maddern_2017_robotCar}, providing a number of weather conditions, such as overcast, rainy, sunny, snowy and night domains.
\vspace{-0.3cm}
\begin{figure}[H]
    \begin{center}
    \includegraphics[width=0.3\linewidth]{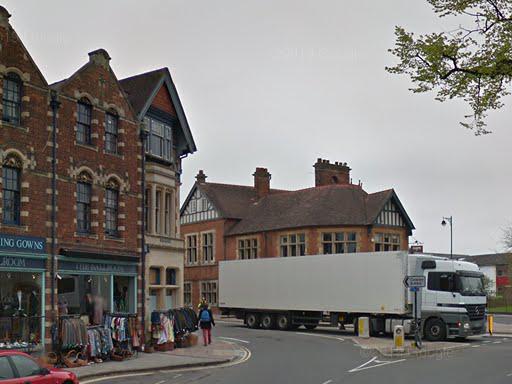}
    \includegraphics[width=0.3\linewidth]{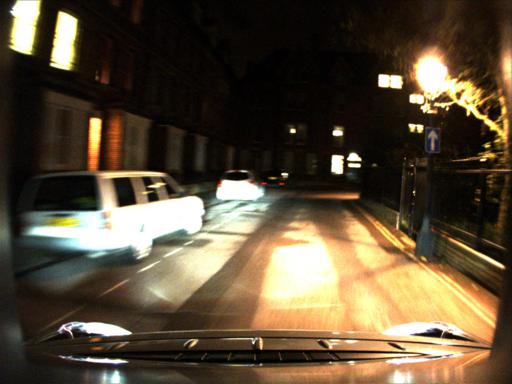}
    \includegraphics[width=0.3\linewidth]{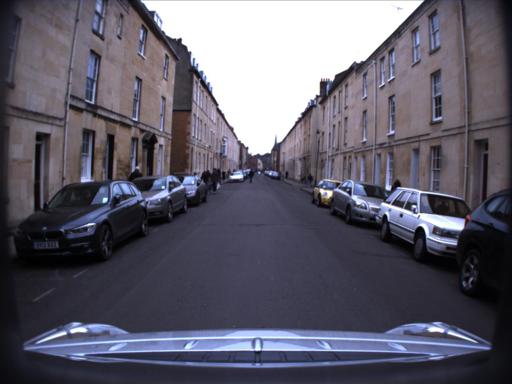}
    \includegraphics[width=0.3\linewidth]{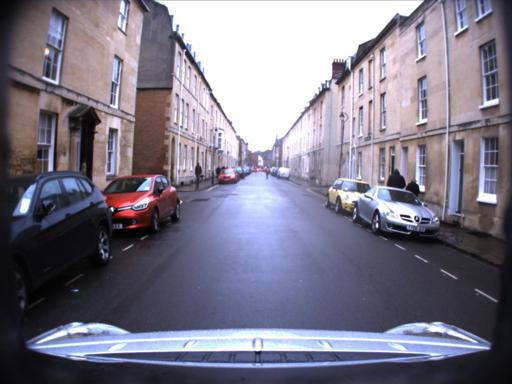}
    \includegraphics[width=0.3\linewidth]{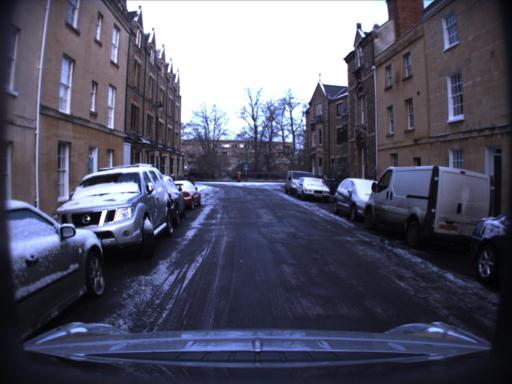}
    \includegraphics[width=0.3\linewidth]{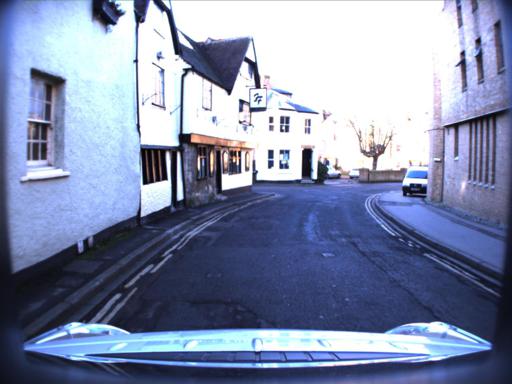}
    \end{center}
    \vspace{-0.6cm}
    \caption{Examples from SVOX: in the top row are an image from the database, and queries from the night and overcast domain; in the bottom row are queries from rain, snow, and sun domains.}
\end{figure}
\vspace{-0.3cm}


\section{Experiments}
\label{sec:supp_experiments}

\begin{table*}
\begin{center}
\begin{adjustbox}{width=\linewidth}
\centering
\begin{tabular}{lccccccccccccccccccc}
\toprule
\multirow{2}{*}{Method} & \multirow{2}{*}{Backbone} & Desc. & \multirow{2}{*}{AmsterTime} & \multirow{2}{*}{Eynsham} & \multirow{2}{*}{Pitts30k} &
\multirow{2}{*}{Pitts250k} & Tokyo & San Francisco & SF-XL & SF-XL \\
& & Dim. & & & & & 24/7 & Landmark & test v1 & test v2 \\
\hline

CosPlace \cite{Berton_2022_cosPlace}     & VGG-16    &  512  & \underline{38.7}/\underline{61.3}/\underline{67.3}/\underline{72.9} & 88.3/92.7/94.1/95.1 & 88.4/94.6/95.7/96.5 & 89.7/96.6/97.8/98.4 & 81.9/90.2/92.4/\underline{95.9} & 80.8/87.5/89.6/91.0 & 65.9/75.3/77.4/80.4 & 83.1/91.3/94.8/95.7\\
\textbf{EigenPlaces (Ours)}              & VGG-16    &  512  & 38.0/59.2/64.8/71.9 & \underline{89.4}/\underline{93.6}/\underline{94.8}/\underline{95.7} & \underline{89.7}/\underline{95.0}/\underline{96.4}/\underline{97.4} & \underline{91.2}/\underline{96.8}/\underline{97.9}/\underline{98.6} & \underline{82.2}/\underline{90.8}/\underline{93.3}/94.3 & \underline{83.8}/\underline{90.6}/\underline{91.8}/\underline{93.0} & \underline{69.4}/\underline{78.4}/\underline{82.0}/\underline{84.8} & \underline{86.3}/\underline{93.6}/\underline{95.3}/\underline{96.2}\\
\hline
NetVLAD \cite{Arandjelovic_2018_netvlad} & VGG-16    & 4096  & 16.3/29.8/36.9/46.4 & \underline{77.7}/\underline{87.8}/\underline{90.5}/\underline{92.5} & 85.0/92.1/94.4/95.9 & 85.9/93.1/95.0/96.3 & 69.8/81.3/82.9/85.7 & 79.1/87.6/89.6/90.8 & 40.0/52.9/57.8/61.9 & 76.9/88.8/91.1/92.8\\
SFRS \cite{Ge_2020_sfrs}                 & VGG-16    & 4096  & \underline{29.7}/\underline{48.5}/\underline{55.6}/\underline{63.4} & 72.3/83.5/87.1/89.8 & \underline{89.1}/\underline{94.6}/\underline{96.1}/\underline{97.0} & \underline{90.4}/\underline{96.3}/\underline{97.6}/\underline{98.2} & \underline{80.3}/\underline{88.6}/\underline{91.7}/\underline{92.7} & \underline{83.1}/\underline{90.0}/\underline{91.8}/\underline{92.8} & \underline{50.3}/\underline{60.0}/\underline{64.9}/\underline{68.5} & \underline{83.8}/\underline{90.5}/\underline{92.8}/\underline{94.3}\\
\hline
CosPlace \cite{Berton_2022_cosPlace}     & ResNet-50 &  128  & \underline{39.9}/\underline{61.8}/\underline{67.9}/\underline{74.2} & 88.6/93.0/94.5/95.4 & 89.0/94.7/96.1/97.1 & 89.6/96.0/97.5/98.3 & \underline{81.0}/\underline{90.8}/\underline{93.7}/94.6 & 82.9/89.6/91.1/91.8 & 69.1/76.5/79.0/82.2 & 86.5/92.6/94.8/\underline{96.7}\\
MixVPR \cite{Alibey_2023_mixvpr}         & ResNet-50 &  128  & 23.1/40.1/49.4/56.9 & 84.8/90.6/92.1/93.4 & 87.7/94.3/95.7/96.9 & 88.7/95.8/97.2/98.2 & 56.8/73.3/80.0/84.1 & 66.9/76.3/80.1/83.3 & 36.7/49.6/55.3/60.1 & 68.4/81.9/87.5/90.6\\
\textbf{EigenPlaces (Ours)}              & ResNet-50 &  128  & 37.9/57.0/65.1/72.9 & \underline{89.1}/\underline{93.7}/\underline{94.8}/\underline{95.8} & \underline{89.6}/\underline{95.6}/\underline{96.7}/\underline{97.3} & \underline{90.2}/\underline{96.4}/\underline{97.7}/\underline{98.4} & 79.4/89.5/\underline{93.7}/\underline{95.6} & \underline{85.5}/\underline{91.5}/\underline{92.5}/\underline{93.3} & \underline{72.4}/\underline{79.4}/\underline{82.3}/\underline{84.5} & \underline{86.6}/\underline{94.3}/\underline{95.3}/\underline{96.7}\\
\hline
CosPlace \cite{Berton_2022_cosPlace}     & ResNet-50 &  512  & \underline{46.4}/67.5/73.3/78.3 & 89.9/93.8/94.8/95.6 & 90.2/95.2/96.3/97.1 & 91.7/97.0/98.1/98.7 & 89.5/94.9/\underline{96.5}/\underline{97.5} & 85.6/90.3/92.3/93.5 & 76.7/82.5/85.6/87.4 & 89.0/95.3/96.3/96.8\\
Conv-AP \cite{Alibey_2022_gsvcities}     & ResNet-50 &  512  & 28.4/46.5/52.8/60.4 & 86.2/91.5/93.1/94.3 & 89.1/94.6/96.1/97.0 & 90.4/96.7/97.8/98.4 & 61.3/77.8/82.5/87.3 & 68.4/78.4/81.6/84.6 & 41.8/53.1/58.0/62.7 & 64.0/74.6/79.1/84.1\\
MixVPR \cite{Alibey_2023_mixvpr}         & ResNet-50 &  512  & 35.8/52.8/60.0/65.9 & 87.6/92.0/93.3/94.3 & 90.4/95.4/96.3/97.2 & 93.0/\underline{97.8}/98.6/\underline{99.0} & 78.4/86.7/90.2/93.0 & 79.4/86.1/88.3/89.6 & 57.7/70.3/74.2/77.4 & 84.3/91.6/94.0/94.5\\
\textbf{EigenPlaces (Ours)}              & ResNet-50 &  512  & 45.7/\underline{68.5}/\underline{74.6}/\underline{80.1} & \underline{90.5}/\underline{94.3}/\underline{95.3}/\underline{96.2} & \underline{91.9}/\underline{96.4}/\underline{97.4}/\underline{97.9} & \underline{93.5}/\underline{97.8}/\underline{98.7}/\underline{99.0} & \underline{89.8}/\underline{95.2}/95.9/96.5 & \underline{89.5}/\textbf{94.5}/\textbf{95.5}/\textbf{96.2} & \underline{82.6}/\underline{87.6}/\underline{90.3}/\underline{91.9} & \underline{90.6}/\underline{95.5}/\textbf{97.2}/\textbf{97.8}\\
\hline
CosPlace \cite{Berton_2022_cosPlace}     & ResNet-50 & 2048  & 47.7/\textbf{69.8}/75.8/81.0 & 90.0/93.9/94.9/95.7 & 90.9/95.7/96.7/97.4 & 92.3/97.4/98.4/98.9 & 87.3/94.0/95.6/97.1 & 87.1/91.1/92.1/92.8 & 76.4/83.3/85.5/88.2 & 88.8/95.0/\underline{96.8}/\underline{97.5}\\
Conv-AP \cite{Alibey_2022_gsvcities}     & ResNet-50 & 2048  & 31.3/49.6/58.1/64.9 & 86.6/91.7/93.1/94.3 & 90.4/95.1/96.4/97.2 & 92.3/97.5/98.4/99.0 & 71.1/81.0/84.8/87.3 & 71.7/81.4/83.9/85.6 & 47.8/58.3/63.1/67.3 & 68.1/80.9/83.9/87.3\\
\textbf{EigenPlaces (Ours)}              & ResNet-50 & 2048  & \textbf{48.9}/69.5/\textbf{76.0}/\textbf{81.4} & \textbf{90.7}/\textbf{94.4}/\textbf{95.4}/\textbf{96.3} & \textbf{92.5}/\textbf{96.8}/\textbf{97.6}/\textbf{98.2} & \textbf{94.1}/\underline{97.9}/\underline{98.7}/\underline{99.1} & \textbf{93.0}/\textbf{96.2}/\textbf{97.5}/\textbf{97.8} & \textbf{89.6}/\underline{94.3}/\underline{95.3}/\underline{95.8} & \textbf{84.1}/\textbf{89.1}/\textbf{90.7}/\textbf{92.6} & \textbf{90.8}/\textbf{95.7}/96.7/\underline{97.5}\\
\hline
Conv-AP \cite{Alibey_2022_gsvcities}     & ResNet-50 & 4096  & 33.9/53.0/59.1/66.8 & 87.5/92.2/93.5/94.6 & 90.5/95.3/\underline{96.6}/\underline{97.5} & 92.3/97.8/98.6/99.0 & 76.2/85.1/87.3/89.2 & 73.7/81.6/84.6/86.3 & 47.5/59.7/63.8/67.8 & 74.4/86.6/89.0/90.8\\
MixVPR \cite{Alibey_2023_mixvpr}         & ResNet-50 & 4096  & \underline{40.2}/\underline{59.1}/\underline{64.6}/\underline{72.5} & \underline{89.4}/\underline{93.2}/\underline{94.3}/\underline{95.1} & \underline{91.5}/\underline{95.5}/96.3/\underline{97.5} & \textbf{94.1}/\textbf{98.2}/\textbf{98.9}/\textbf{99.3} & \underline{85.1}/\underline{91.7}/\underline{94.3}/\underline{95.6} & \underline{83.8}/\underline{90.3}/\underline{91.1}/\underline{92.5} & \underline{71.1}/\underline{78.2}/\underline{79.7}/\underline{82.3} & \underline{88.5}/\underline{93.6}/\underline{94.5}/\underline{96.0}\\
Conv-AP \cite{Alibey_2022_gsvcities}     & ResNet-50 & 8192  & 35.0/53.8/60.9/68.2 & 87.6/92.4/93.6/94.5 & 90.5/95.2/96.4/97.3 & 92.6/97.5/98.4/99.0 & 72.1/84.1/87.6/90.5 & 74.4/82.9/85.5/87.8 & 49.3/61.0/64.8/69.8 & 75.8/85.1/89.0/91.3\\

\bottomrule
\end{tabular}
\end{adjustbox}
\end{center}
\caption{\textbf{Recalls (R@1 / R@5 / R@10 / R@20) on multi-view datasets}, split according to the utilized backbone and descriptors dimension. Best overall results on each dataset are in \textbf{bold}, best results for each group are \underline{underlined}.}
\label{tab:full_multi_view}
\end{table*}

\begin{table*}
\begin{center}
\begin{adjustbox}{width=\linewidth}
\centering
\begin{tabular}{lccccccccccccccccccc}
\toprule
\multirow{2}{*}{Method} & \multirow{2}{*}{Backbone} & Desc. & MSLS &
\multirow{2}{*}{Nordland} & \multirow{2}{*}{St Lucia} & SVOX & SVOX & SVOX & SVOX & SVOX \\
& & Dim. & Val & & & Night & Overcast & Rain & Snow & Sun \\
\hline

CosPlace \cite{Berton_2022_cosPlace}     & VGG-16    &  512  & 82.6/89.9/\underline{92.0}/\underline{94.3} & \underline{58.5}/\underline{73.7}/\underline{79.4}/\underline{84.8} & 95.3/97.9/98.9/99.5 & \underline{44.8}/\underline{63.5}/\underline{70.0}/\underline{77.6} & 88.5/93.9/95.2/96.7 & \underline{85.2}/\underline{91.7}/\underline{93.8}/\underline{95.3} & 89.0/94.0/94.6/96.0 & 67.3/79.2/83.8/88.4\\
\textbf{EigenPlaces (Ours)}              & VGG-16    &  512  & \underline{84.2}/\underline{90.0}/91.8/94.1 & 54.5/70.1/76.4/82.4 & \underline{95.4}/\underline{98.1}/\underline{99.5}/\underline{99.7} & 42.3/61.0/68.5/75.8 & \underline{89.4}/\underline{94.4}/\underline{95.6}/\underline{97.4} & 83.5/91.6/92.8/94.6 & \underline{89.2}/\underline{94.4}/\underline{95.5}/\underline{96.1} & \underline{69.7}/\underline{82.2}/\underline{86.1}/\underline{89.8}\\
\hline
NetVLAD \cite{Arandjelovic_2018_netvlad} & VGG-16    & 4096  & 58.9/70.8/75.0/79.1 & 13.1/21.1/26.1/32.0 & 64.6/80.3/85.8/91.3 & 8.0/17.4/23.1/29.6 & 66.4/81.5/85.7/89.3 & 51.5/69.3/74.7/80.4 & 54.4/71.8/77.2/82.4 & 35.4/52.7/58.8/65.8\\
SFRS \cite{Ge_2020_sfrs}                 & VGG-16    & 4096  & \underline{70.0}/\underline{80.0}/\underline{83.5}/\underline{86.1} & \underline{16.0}/\underline{24.1}/\underline{28.7}/\underline{34.4} & \underline{75.9}/\underline{86.6}/\underline{91.2}/\underline{94.3} & \underline{28.6}/\underline{40.6}/\underline{46.4}/\underline{52.1} & \underline{81.1}/\underline{88.4}/\underline{91.2}/\underline{92.9} & \underline{69.7}/\underline{81.5}/\underline{84.6}/\underline{87.7} & \underline{76.0}/\underline{86.1}/\underline{89.4}/\underline{91.6} & \underline{54.8}/\underline{68.3}/\underline{74.1}/\underline{78.5}\\
\hline
CosPlace \cite{Berton_2022_cosPlace}     & ResNet-50 &  128  & \underline{85.5}/\underline{92.3}/93.2/94.6 & \underline{54.7}/\underline{70.9}/\underline{77.9}/\underline{83.4} & 98.7/99.8/\underline{99.9}/\textbf{100.0} & \underline{35.4}/\underline{55.4}/\underline{63.8}/\underline{71.0} & 88.5/96.0/96.9/97.5 & 80.4/90.3/94.1/95.9 & 86.6/95.1/96.4/97.4 & 65.2/80.3/84.4/88.4\\
MixVPR \cite{Alibey_2023_mixvpr}         & ResNet-50 &  128  & 79.1/87.4/90.3/92.0 & 47.8/66.5/73.9/80.5 & \underline{99.0}/\textbf{99.9}/\underline{99.9}/99.9 & 25.9/43.3/50.9/59.2 & \underline{92.3}/\underline{96.6}/97.4/97.7 & 80.9/91.2/93.8/94.9 & 87.7/94.6/95.6/96.9 & \underline{73.5}/\underline{88.1}/\underline{91.2}/\underline{94.3}\\
\textbf{EigenPlaces (Ours)}              & ResNet-50 &  128  & 83.4/90.9/\underline{93.5}/\underline{95.1} & 50.5/66.8/73.6/80.0 & 98.8/99.7/\underline{99.9}/\textbf{100.0} & 29.0/48.5/57.7/65.4 & 90.9/96.2/\underline{97.6}/\underline{98.3} & \underline{83.8}/\underline{92.8}/\underline{94.6}/\underline{96.7} & \underline{91.1}/\underline{97.0}/\underline{97.9}/\underline{99.0} & 68.5/83.7/88.2/91.8\\
\hline
CosPlace \cite{Berton_2022_cosPlace}     & ResNet-50 &  512  & 86.9/93.2/94.2/95.5 & 66.5/79.7/84.8/88.9 & 99.1/\textbf{99.9}/\textbf{100.0}/\textbf{100.0} & \underline{51.6}/68.8/76.1/80.9 & 90.0/96.6/97.2/97.6 & 87.3/94.7/95.7/97.3 & 89.5/97.0/98.0/98.2 & 75.9/88.3/92.2/94.6\\
Conv-AP \cite{Alibey_2022_gsvcities}     & ResNet-50 &  512  & 82.3/90.3/91.6/93.5 & 59.2/74.6/80.1/85.2 & 99.2/\textbf{99.9}/99.9/99.9 & 36.0/52.5/61.2/67.9 & 90.5/95.9/96.9/98.2 & 80.3/90.0/93.0/95.4 & 86.4/95.3/96.6/98.3 & 75.3/88.1/91.5/93.1\\
MixVPR \cite{Alibey_2023_mixvpr}         & ResNet-50 &  512  & 83.6/91.5/93.4/94.3 & 67.2/81.0/\underline{85.9}/\underline{90.0} & 99.2/\textbf{99.9}/\textbf{100.0}/\textbf{100.0} & 44.8/63.2/71.0/77.0 & \underline{93.9}/\underline{97.7}/\underline{98.3}/\underline{98.7} & 86.4/93.9/96.3/97.4 & \underline{93.9}/\underline{97.6}/97.9/98.5 & 78.7/91.2/93.6/95.4\\
\textbf{EigenPlaces (Ours)}              & ResNet-50 &  512  & \textbf{89.5}/\underline{93.6}/\underline{94.5}/\underline{96.1} & \underline{67.9}/\underline{81.1}/85.6/89.6 & \underline{99.5}/\textbf{99.9}/\textbf{100.0}/\textbf{100.0} & 51.5/\underline{70.8}/\underline{78.4}/\underline{84.0} & 92.8/97.6/97.9/98.4 & \underline{89.0}/\underline{95.5}/\underline{97.1}/\underline{98.1} & 92.0/97.5/\underline{98.3}/\underline{98.7} & \underline{83.1}/\underline{93.8}/\underline{95.7}/\textbf{97.1}\\
\hline
CosPlace \cite{Berton_2022_cosPlace}     & ResNet-50 & 2048  & 87.4/\textbf{94.1}/94.9/95.9 & \underline{71.9}/\underline{83.8}/\underline{88.1}/91.5 & \underline{99.6}/\textbf{99.9}/\textbf{100.0}/\textbf{100.0} & 50.7/67.4/74.8/80.2 & 92.2/97.7/97.9/\underline{98.7} & 87.0/95.1/96.8/97.5 & 92.0/\textbf{98.4}/\textbf{98.9}/\textbf{99.1} & 78.5/89.7/93.1/94.8\\
Conv-AP \cite{Alibey_2022_gsvcities}     & ResNet-50 & 2048  & 81.2/89.5/91.6/93.6 & 62.3/76.9/82.0/86.7 & 99.3/\textbf{99.9}/\textbf{100.0}/\textbf{100.0} & 37.9/57.1/65.4/72.8 & 92.0/96.1/97.2/98.5 & 83.7/93.4/95.2/97.2 & 90.2/95.7/97.5/98.4 & 80.3/90.5/93.8/95.4\\
\textbf{EigenPlaces (Ours)}              & ResNet-50 & 2048  & \underline{89.1}/93.8/\textbf{95.0}/\textbf{96.2} & 71.2/\underline{83.8}/\underline{88.1}/\underline{91.6} & \underline{99.6}/\textbf{99.9}/\textbf{100.0}/\textbf{100.0} & \underline{58.9}/\underline{76.9}/\underline{82.6}/\underline{87.0} & \underline{93.1}/\underline{97.8}/\underline{98.3}/\underline{98.7} & \underline{90.0}/\underline{96.4}/\underline{98.0}/\textbf{98.5} & \underline{93.1}/97.6/98.2/98.6 & \textbf{86.4}/\textbf{95.0}/\textbf{96.4}/\underline{96.8}\\
\hline
Conv-AP \cite{Alibey_2022_gsvcities}     & ResNet-50 & 4096  & 82.8/89.9/91.8/94.5 & 59.6/74.4/79.7/84.9 & 99.6/\textbf{99.9}/\textbf{100.0}/\textbf{100.0} & 41.9/61.4/68.7/76.5 & 91.2/95.8/97.1/98.1 & 81.9/92.6/95.2/96.9 & 87.9/95.7/97.7/98.7 & 82.0/91.7/94.5/\underline{96.0}\\
MixVPR \cite{Alibey_2023_mixvpr}         & ResNet-50 & 4096  & \underline{87.2}/\underline{93.1}/\underline{94.3}/\underline{95.4} & \textbf{76.2}/\textbf{86.9}/\textbf{90.3}/\textbf{93.3} & 99.6/\textbf{99.9}/\textbf{100.0}/\textbf{100.0} & \textbf{64.4}/\textbf{79.2}/\textbf{83.1}/\textbf{87.7} & \textbf{96.2}/\textbf{98.3}/\textbf{98.9}/\textbf{99.2} & \textbf{91.5}/\textbf{97.2}/\textbf{98.1}/\textbf{98.5} & \textbf{96.8}/\textbf{98.4}/\textbf{98.9}/\underline{99.0} & \underline{84.8}/\underline{93.2}/\underline{94.7}/95.9\\
Conv-AP \cite{Alibey_2022_gsvcities}     & ResNet-50 & 8192  & 82.4/90.4/92.0/94.3 & 62.9/77.3/82.5/86.8 & \textbf{99.7}/\textbf{99.9}/99.9/99.9 & 43.4/63.1/71.6/79.1 & 91.9/96.6/98.3/98.6 & 82.8/93.0/95.6/96.1 & 91.0/96.7/97.6/98.4 & 80.4/90.3/93.2/95.0\\

\bottomrule
\end{tabular}
\end{adjustbox}
\end{center}
\caption{\textbf{Recalls (R@1 / R@5 / R@10 / R@20) on frontal-view datasets}, split according to the utilized backbone and descriptors dimension. Best overall results on each dataset are in \textbf{bold}, best results for each group are \underline{underlined}.}
\label{tab:full_frontal_view}
\end{table*}

\subsection{Further results}
In this section we report different values of recalls for the same datasets as in the main paper.
Results on multi-view datasets are in \cref{tab:full_multi_view} , whereas on frontal-view datasets in \cref{tab:full_frontal_view}.

\subsection{Qualitative results}
\begin{figure*}
    \begin{center}
    \includegraphics[width=0.0353\textwidth]{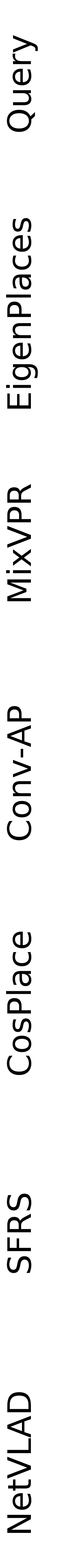}
    \includegraphics[width=0.12\textwidth]{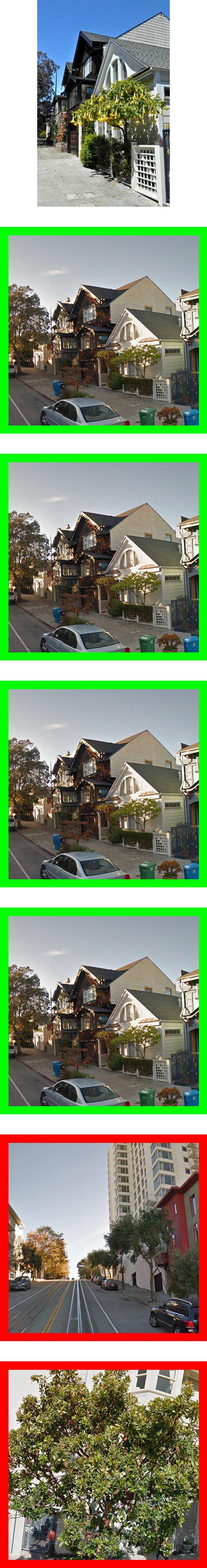}
    \includegraphics[width=0.12\textwidth]{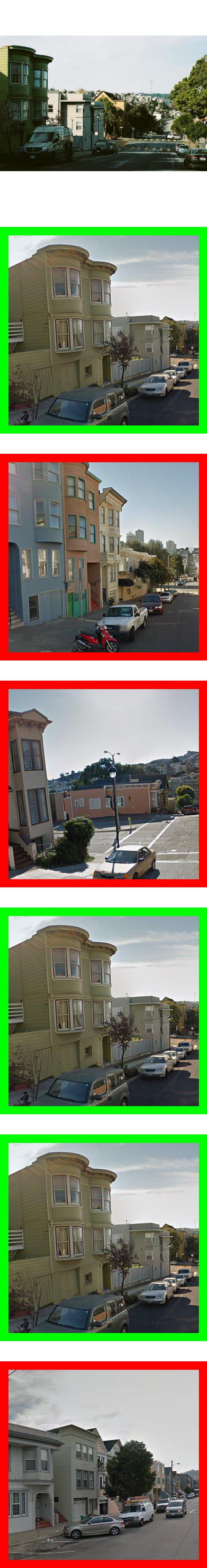}
    \includegraphics[width=0.12\textwidth]{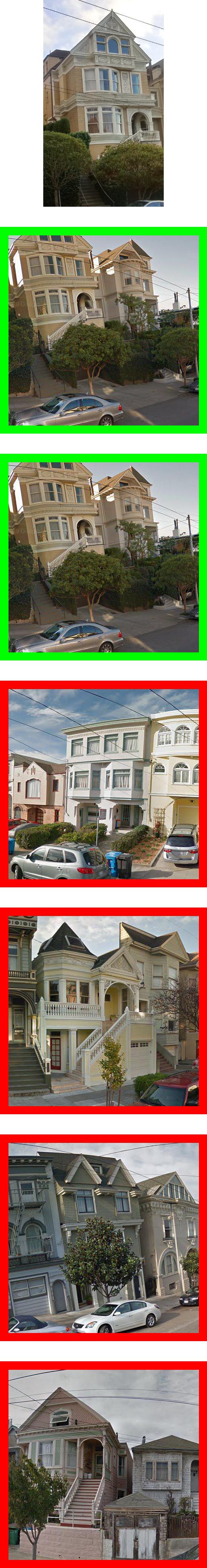}
    \includegraphics[width=0.12\textwidth]{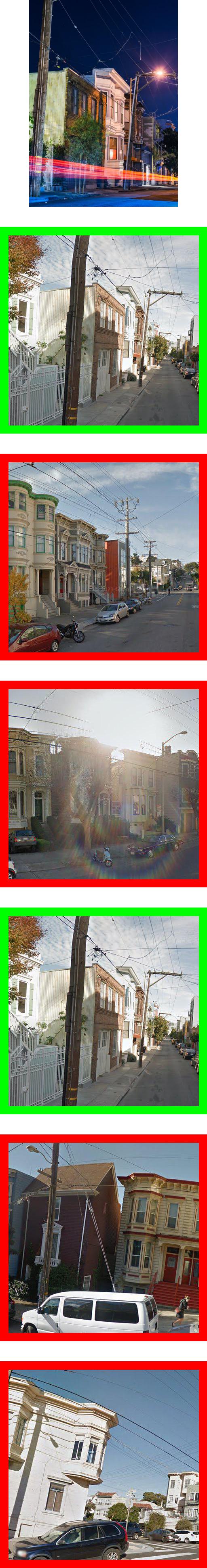}
    \includegraphics[width=0.12\textwidth]{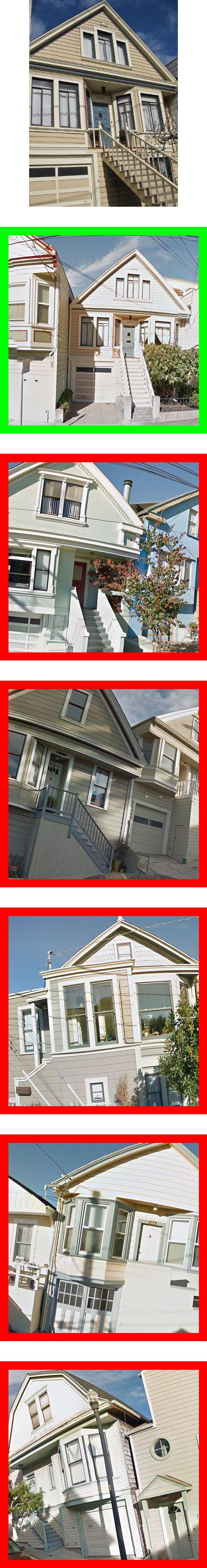}
    \includegraphics[width=0.12\textwidth]{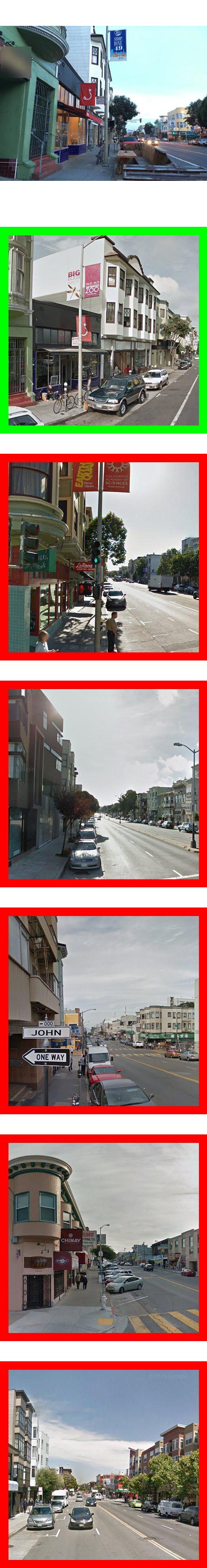}
    \includegraphics[width=0.12\textwidth]{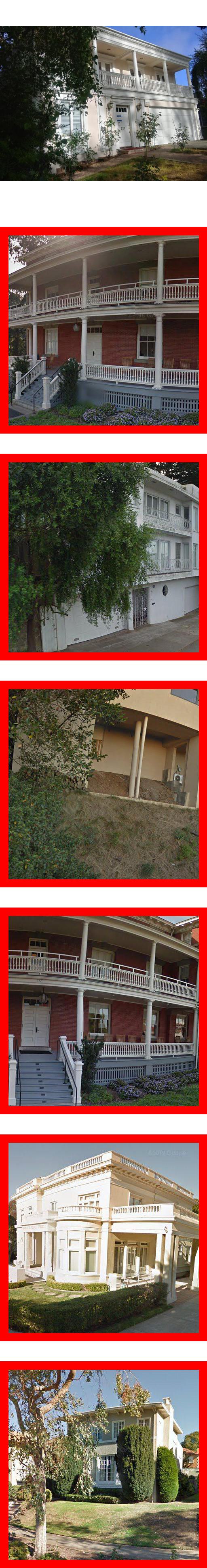}
    \end{center}
    \caption{\textbf{Qualitative results with most popular methods.} Each column represents a query (top row) and the first predicted image from the database. We can see that EigenPlaces is able to better handle challenging viewpoints than previous methods.}
    \label{fig:qualitatives}
\end{figure*}

Some qualitative results are shown in \cref{fig:qualitatives}.
The figure allows to understand the strengths of EigenPlaces in a more visual and intuitive way. We can see that EigenPlaces is able to handle difficult points of view, such as photos taken from the sidewalk.

\end{document}